# Autonomous Droplet Navigation via Model-Based Reinforcement Learning

Rajneesh Anand[1] and Mayuresh V. Kothare[1,*]
[1]*Department of Chemical and Biomolecular Engineering, Lehigh University, Bethlehem, PA 18015, USA.*
*Corresponding author. E-mail: mvk2@lehigh.edu

**Abstract**

Precise manipulation of liquid droplets underpins lab-on-a-chip platforms for diagnostics, chemical synthesis, and biological assays. Yet autonomous droplet transport through confined geometries of varying complexity remains an open challenge. Droplets exhibit contact-angle hysteresis, deformability, and capillary pinning, which make their response to actuation nonlinear and history dependent, that classical controllers and pre-programmed trajectories cannot cope in multi-turn environments. Here we demonstrate autonomous navigation of a liquid droplet through geometries of increasing complexity on a gravity driven (Labyrinth) platform using model-based reinforcement learning. A thin silicone oil film reduces contact-line pinning while two-axis tilt supplies the gravitational driving force, and an overhead camera tracks the droplet in real time. An offline-trained policy discovers effective tilt strategies from limited physical interaction data, without simulation or analytical droplet models. The system operates under partial observability, as oil-film thickness, instantaneous contact angle, and droplet deformation state remain hidden from the controller. Despite these challenges, the learned policy achieves reliable navigation across straight, right-angle, and curved-arc paths, including outside-corner geometries. We further demonstrate that a policy trained on a simpler geometry transfers to complex ones, succeeding zero-shot on right-angle and staircase paths and reaching full success on a curved arc with a fifth of the training data. The findings suggest promising avenues for enabling droplet based microfluidic systems to serve as intelligent chemical laboratories.

## Introduction

The transport and manipulation of discrete liquid droplets across solid surfaces is a fundamental problem in chemical engineering with far-reaching implications for nanoliter reaction vessels in lab-on-a-chip diagnostics, multistep synthesis and high-throughput screening with further applications in self-cleaning surfaces and thermal management **[1–3]**. Over the past two decades, a rich body of work has established that droplets can be set into motion by imposing spatial gradients in surface energy **[4]**, vibration **[7-10]**, Electrowetting-on-dielectric (EWOD) **[5, 6],** or combinations thereof. Despite this diversity of physical mechanisms, a persistent barrier to robust droplet transport remains contact angle hysteresis, which resists the onset of motion **[11, 12]**. Hysteresis arises from chemical heterogeneity, surface roughness, and adsorption at the solid–liquid interface, and its magnitude varies unpredictably with surface preparation, ambient conditions, and droplet history **[11]**. One widely adopted strategy to mitigate hysteresis is to coat the substrate with a thin lubricant film, the principle underlying slippery liquid-infused porous surfaces (SLIPS) **[13]** and oil-infused surfaces **[14, 15]**, which replaces the solid–liquid contact line with a liquid–liquid interface exhibiting near-zero hysteresis. Nevertheless, even with lubrication, capillary trapping at geometric constrictions such as wall boundaries, corners, channel narrowing introduces spatially heterogeneous resistance that cannot be overcome by simple feedforward strategies **[16, 17]**.

Despite decades of progress in understanding and engineering droplet transport mechanisms, all existing experimental demonstrations rely on either preprogrammed actuation sequences (open loop) or manual human control, neither of which is robust. In EWOD systems, electrode activation patterns are scripted a priori and executed sequentially **[5, 6]**; in vibration-based platforms, the operator manually tunes frequency and amplitude **[7-10]**. Moreover, these manually selected parameters are fixed and geometry-specific, with no capacity to adapt mid-path resistance or guarantee optimality. To date, no prior study has demonstrated closed-loop autonomous navigation of a free liquid droplet along an arbitrary path on an open (unconfined) surface whether by classical feedback control (e.g., proportional-integral-derivative, PID) or by any learning-based method. This absence of autonomy poses a concrete barrier to scalability. As droplet microfluidic platforms scale toward complex multi-step protocols with branching paths, real-time decision-making at each junction becomes essential, and human operators cannot keep pace with the millisecond timescales of capillary dynamics **[18, 19]**. The absence of autonomous droplet navigation stands in sharp contrast to the rapid progress in autonomous control of other micro-scale agents, including magnetic microrobots **[20, 21]**, ultrasound-driven microbubble swarms **[22]**, and colloidal swimmers **[23]**, all of which have benefited from recent advances in machine learning.

Reinforcement learning (RL) has established itself as a transformative paradigm for sequential decision-making in environments with complex, partially known dynamics **[24]**. Foundational demonstrations include superhuman gameplay **[25-27]**, dexterous robotic manipulation **[28, 29]**, and autonomous drone racing at champion level **[30]**. Closer to engineering practice, RL has been deployed for closed-loop control of physical processes whose dynamics are nonlinear, including closed-loop regulation of cardiovascular system with vagus nerve stimulation **[31]** and process control in tokamak plasma confinement **[32]**. In the micro-scale domain, RL has been applied to navigate self-thermophoretic microswimmers under Brownian noise **[23]**, to steer magnetic microrobot swarms through deep learning-based distribution planning **[21]**. These studies collectively establish RL and model-based reinforcement learning (MBRL) in particular – as a powerful approach for autonomous micro-scale navigation. However, every existing RL-controlled micro-agent is a solid object (a magnetic bead, a rigid micromachine) whose shape, mass, and contact mechanics are fixed. A liquid droplet, by contrast, is deformable, susceptible to breakup, and governed by interfacial thermodynamics (contact angle, surface tension) that introduce fundamentally different and richer nonlinear dynamics **[11, 33, 34]**. Autonomous control of a free liquid droplet on an open surface by reinforcement learning remains entirely unexplored. Droplet decision-making can also emerge from physicochemical mechanisms alone **[43]**, whereas we actuate externally and learn control from visual feedback. Among RL paradigms, MBRL offers compelling advantages for physical micro-systems where data collection is slow, expensive, and operationally constrained **[35-39]**. Model-free algorithms (PPO (proximal policy optimization) **[40]**, SAC (soft actor-critic) **[41]**) require millions of environment interactions to converge. In our setting, the system is partially observable, nonlinear, and expensive to simulate from first principles (requiring resolution of the Navier–Stokes equations coupled with dynamic contact-line models **[34, 42]**), making it naturally suited to learned policy that capture the essential dynamics directly from visual observations.

Here, we introduce DropletRunner, the first demonstration to date of closed-loop autonomous liquid droplet navigation on an open, unconfined surface via MBRL, a new problem domain at the intersection of droplet microfluidics and reinforcement learning. We enable autonomy in droplet navigation from a start position to a goal along prescribed paths of increasing geometric complexity: a straight-line (I-shape), a 90° turn navigating the inside wall (L-in), the outside wall without confinement (L-out), a continuous-curvature arc navigating the inside wall (Arc-in), and the outside wall (Arc-out). **Fig. 1e** shows all these five trajectories. The agent achieves 100%, 95%, 90%, 100%, and 90% success on the I-, L-in-, L-out-, Arc-in-, and Arc-out-shape respectively, without any prior knowledge of the surface physics and oil properties. Three substantial findings carry this work beyond a navigation demonstration. First, we show that a PID controller on the same hardware with geometry-specific tuned gains achieves no more than 20% success. Second, the learned policy reveals that the agent has autonomously discovered an emergent oscillatory control strategy. It applies alternating-polarity current commands when the droplet becomes trapped at a geometric constriction. This is the first instance of an RL agent independently rediscovering a known physical droplet transport mechanism without any encoding of the underlying physics. It suggests that MBRL may serve as a tool for discovering novel control strategies in chemical engineering systems more broadly. Third, we demonstrate a zero-shot transfer learning framework in which a policy trained on the simpler (I-) shape navigates the droplet on the L-in-, Arc-in-, and a new Staircase-shape that introduces greater complexity and uncertainty by combining an inside wall turn and an outside wall turn without confinement. The Stair-shape, built from a repeating unit cell, serves as proof of concept for infinite-horizon autonomous droplet transport along arbitrarily extended serpentine paths.

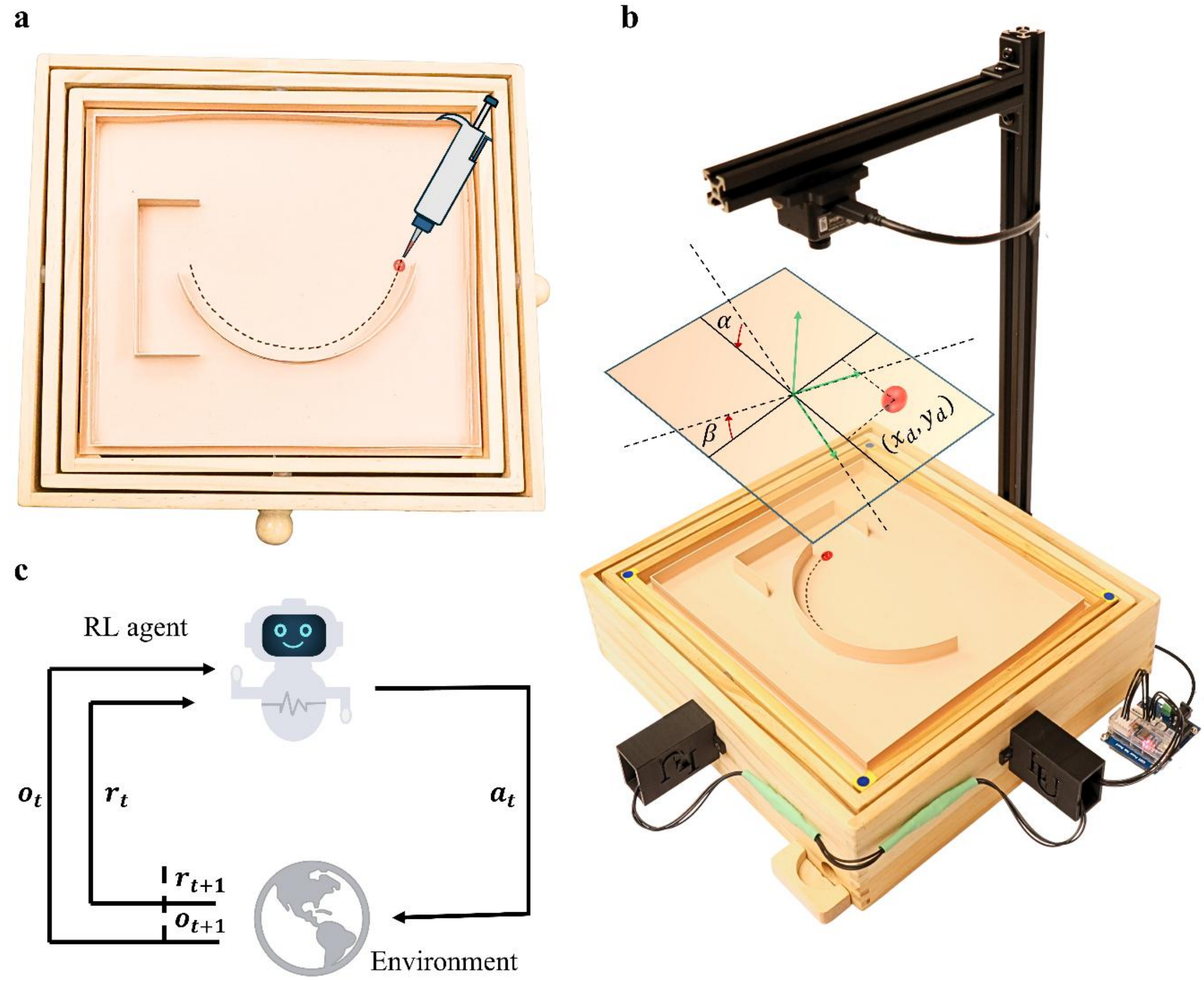

**d**

**Guidelines for autonomous droplet manipulation via MBRL**

| **Step 1:** Formulate droplet manipulation as a partially observable RL problem with progress-based rewards. | **Step 2:** Learn environment dynamics from physical episodes – no simulation, no physics model. | **Step 3:** Deploy the learned policy to autonomously navigate complex path in the physical environment. |
|---|---|---|

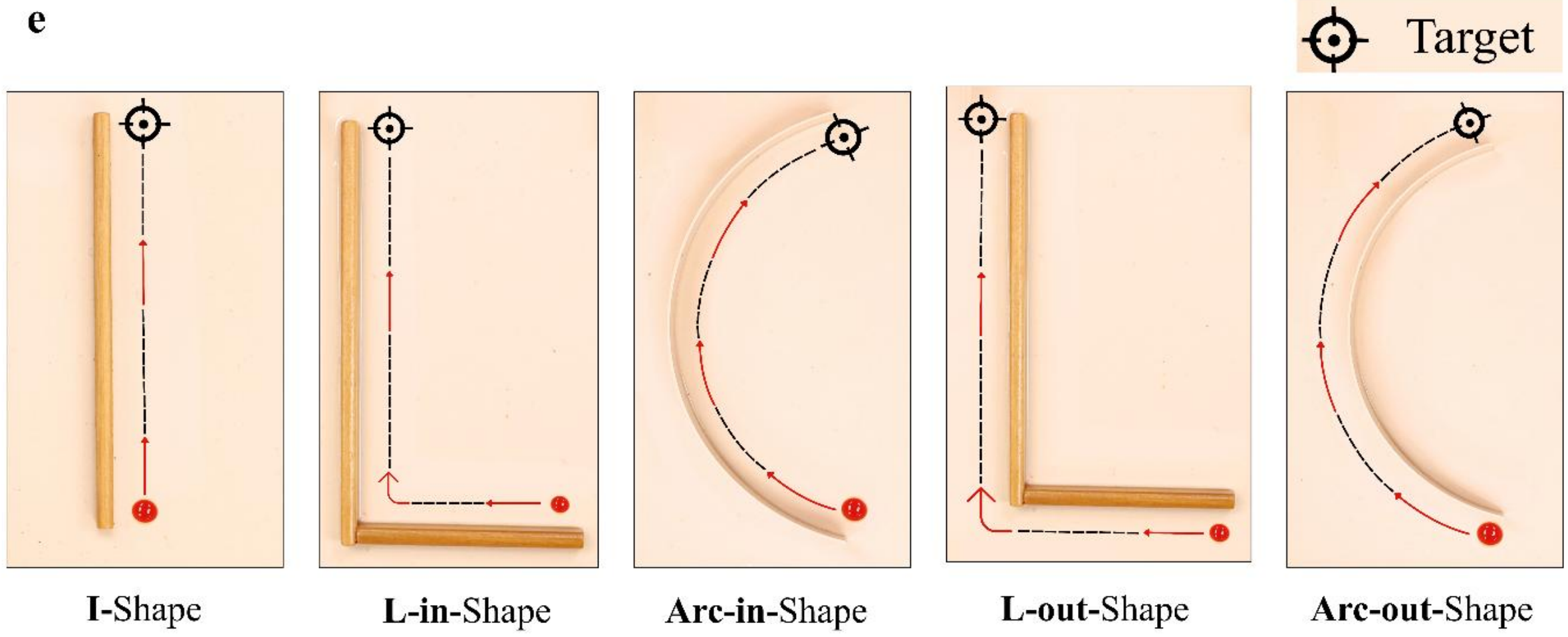

**Fig. 1 | Autonomous droplet navigation on an oil-lubricated surface using MBRL. a,** The Labyrinth board showing a 3D-printed PLA channel geometry coated with a thin silicone oil

film. A red-dyed water droplet navigates the channels via manual tilting by a human operator. **b,** The same board integrated into the autonomous MBRL platform. An overhead camera provides visual feedback, two Dynamixel motors tilt the board, and a U2D2 communication interface connects the motors to the control laptop. Together, these components transform the manual Labyrinth into a closed-loop system in which the RL agent observes, decides, and actuates without human intervention. Overlay defines tilt angles $\alpha$ and $\beta$ and droplet position $(x_d, y_d)$. **c,** High-level RL formulation. At each time step, the agent receives a partial observation $o_t$ (camera image and state vector) and a reward $r_t$ (progress along the path), and outputs a continuous action $a_t$ specifying the two motor currents. **d,** Three-step guidelines for autonomous droplet manipulation via MBRL. **e,** Five trajectories of increasing complexity: I-shape (straight), L-in-shape (corner), Arc-in-shape (continuous curvature), L-out-shape (corner with exterior wall), and Arc-out-shape (exterior wall). Red circles mark the start position; crosshair symbols mark the target. Dashed arrows indicate the navigation direction.

## Problem formulation

We formulate autonomous droplet navigation as a sequential decision-making problem in which an RL agent learns to control the two-axis tilt of the board to transport a liquid droplet from a start position to a goal along a prescribed path (**Fig. 1a**). A water droplet sits on a thin film of oil-lubricated PLA board (**Fig. 1a**) and is actuated by gravity balanced by viscous drag through two-axis motor-driven tilting (**Fig. 1b**). At each discrete time step ($t$), the agent observes the environment state and selects an action; this process repeats until the episode terminates. The policy $\pi$ (a neural network) maps observations to actions and is updated through training to maximize cumulative reward. The observation space ($o$) comprises a visual component and a state vector (**Eq. 1**), as shown in **Fig. 2a**.

$$o_t = [I_t, v_t] \tag{1}$$

where $I_t \in \mathbb{R}^{64\times64\times3}$ is a downsampled RGB image from the overhead camera estimating the current board (see **Supplementary Information Note S1**) states **[44]** (**Fig. 1b**), and $v_t \in \mathbb{R}^{14}$ is a state vector containing the droplet's estimated position $(x_d, y_d)$, board tilt angle $(\alpha, \beta)$, and five future waypoint positions along the path, each with $x_d$ and $y_d$ coordinates. The control action is applied to both motors simultaneously. The agent evaluates (**Fig. 2b, c**) its performance using a reward signal $r$ generated by the environment (**Fig. 1c**) . We define the reward

$$r_t = \Delta p_t - 0.1 \times d_t + r_{\text{terminal}} \tag{2}$$

where $\Delta p_t = p_t - p_{t-1}$ is the incremental progress, $d_t$ is the perpendicular distance from the droplet to the nearest path segment, and $r_{terminal}$ is a sparse terminal signal: a +10.0 bonus is added to the step reward when it reaches to the goal and a penalty of −1.0 if it deviates from the path and the episode terminates. The optimal policy $\pi^*$ (**Eq. 3**) maximizes the expected cumulative ($\mathbb{E}$) discounted reward.

$$\pi^* = \arg\max_{\pi} \mathbb{E}\left[\sum_{t=0}^{T} \gamma^t \; r_t\right] \tag{3}$$

where $\gamma$ (< 1) is the discount factor and T is the episode length. An episode terminates when any of three conditions is met: (i) the droplet enters the goal region (success), (ii) the droplet's lateral deviation from the path exceeds a threshold $d_{max}$ (failure), or (iii) the number of time steps reaches the maximum episode length 600, corresponding to a 30-second timeout (failure).

We employ DreamerV3 **[35]**, an MBRL algorithm that learns a latent-space world model from collected experience. The algorithm consists of a world model that learns the environment dynamics, an actor that selects actions, and a critic that estimates expected returns, as shown in **Fig. 2b, c**. The environment is partially observable as the overhead camera provides the droplet's position and the board's tilt angle, but cannot resolve the oil film thickness, the local contact angle, the meniscus geometry at wall boundaries, all of which govern the droplet's response to tilt commands. Two physically distinct situations (for example, fresh oil versus depleted oil at the same board position) produce nearly identical observations yet require different control strategies. To address this partial observability, DreamerV3 **[35]** employs a recurrent state-space model (RSSM), shown in **Fig. 2b,** that maintains a two-part latent state ($s_t = (h_t, z_t)$) at each time step. $h_t$ is the deterministic recurrent state, a fixed-size hidden vector of a gated recurrent unit (GRU) that accumulates a compressed summary of the entire episode history (observations and actions from steps 0 through $t-1$); and $z_t$ is the stochastic latent state, a set of categorical random variables sampled from a learned distribution that captures the agent's belief about the unobserved physics (effective viscosity, local adhesion strength, motor response characteristics). A detailed training procedure is discussed in **Supplementary Information Note S1.**

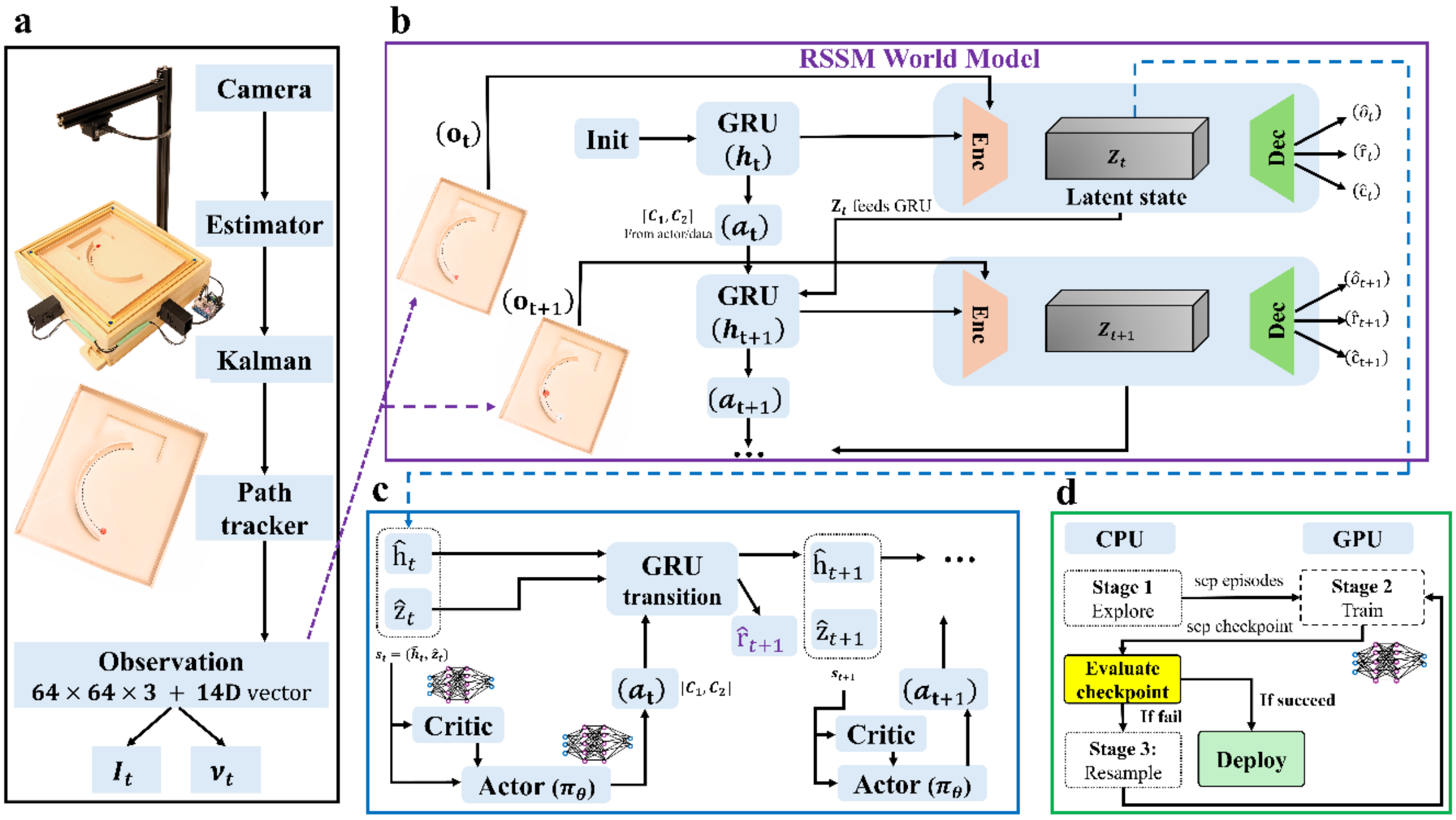

**Fig. 2 | MBRL framework for autonomous liquid droplet navigation. a**, An overhead camera captures raw frames of the tilting board. The estimator extracts droplet and board features, the Kalman filter suppresses glare-induced tracking dropouts, and the path tracker computes progress and deviation relative to the prescribed waypoint sequence. The resulting observation $o_t$ comprises an RGB image ($I_t$) and a 14D state vector ($v_t$). **b**, Recurrent state-space model (RSSM) world model, trained on real episodes. At each time step, the GRU (Gated recurrent network) integrates the previous latent state $z_t$ and the executed (current) control

action $a_t = [C_1, C_2]$ to update the deterministic hidden state $h_{t+1}$. The encoder fuses the new observation $o_{t+1}$ with $h_{t+1}$ to produce the posterior stochastic state $z_{t+1}$, capturing unobservable physical quantities. Decoder heads reconstruct the observation ($\hat{o}_t$), reward ($\hat{r}_t$), and episode continuation ($\hat{c}_t$) from the joint latent state. **c**, The same GRU from **b** is unrolled forward without real observations. The actor $\pi_\theta$ proposes actions and the critic estimates returns, both trained entirely on imagined rollouts generated by the world model. **d**, Stage 1: random exploration episodes are collected and transferred for training. Stage 2: the world model, actor, and critic are jointly trained. Stage 3: the resulting policy is evaluated on the physical system; if the policy fails, guided exploration episodes are collected under the learned policy with exploration noise and training resumes from the existing checkpoint. Upon convergence, the final policy is deployed for autonomous evaluation.

**Physical environment**

The physical environment consists of a water droplet (volume approximately 20-30 μL, dyed red for visual tracking) resting on a thin film of AR 20 silicone oil that coats a flat PLA board. The oil film replaces the solid-liquid contact line with a liquid-liquid interface, mitigating contact angle hysteresis **[14, 15]**, and provides a low-friction substrate that permits droplet transport at small tilt angles **(Movie S1)**. In modified platform (**Fig. 1b**), an overhead camera provides real-time visual feedback. When the board tilts by an angle $\theta$ from the horizontal, the droplet experiences a gravitational body force along the tilted surface.

$$F_g = \rho V g \sin\theta \tag{4}$$

where $\rho$ is the droplet density, $V$ is the droplet volume, and $g$ is the gravitational acceleration. This driving force is balanced by viscous drag from the underlying oil film. Under the lubrication approximation, the drag force (**Eq. 5**) on a droplet translating at velocity $U$ over a thin oil film of viscosity $\mu$ and thickness $h_0$ scales as **[42]**

$$F_d \sim \frac{\mu U R^2}{h_0} \tag{5}$$

where $R$ is the droplet footprint radius. At steady state, the force balance $F_g = F_d$ yields the characteristic droplet velocity (**Eq. 6**).

$$U \sim \frac{\rho V g h_0 \sin\theta}{\mu R^2} \tag{6}$$

Since the Dynamixel motors operate in current control mode, the applied current is proportional to the motor torque, which sets the tilt angle of the board (**see Supplementary Information, Note S1**). The normalized action sets the motor current, which generates a proportional torque on the tilt axis, inclining the board at angle θ, which in turn establishes the gravitational driving force that propels the droplet at a velocity governed by the local force balance.

In addition to the gravitational driving force and viscous drag, a third force arises from the meniscus that forms at the contact line between the droplet, the oil film, and the surrounding air. When the droplet is near a physical wall boundary, the meniscus deforms asymmetrically, generating a capillary adhesion force that resists motion away from the wall **[45, 46]**. At geometric constrictions such as the L-shape corner, the droplet must overcome this adhesion barrier to change direction. The magnitude of the capillary adhesion depends on the local wall geometry, oil film thickness, and droplet size, and varies spatially along the path that is not directly observable from the overhead camera image. This spatially heterogeneous resistance is the primary source of navigation difficulty. The oil film thickness $h_0$ sets the viscous drag and therefore the droplet's velocity response to tilt action, and it varies across the board surface and decreases over time as repeated droplet passages displace the oil.

## Results

### Comparison of MBRL and PID

We compared our MBRL policy against a manually tuned PID controller ($K_p = 4, K_i = 0.1, K_d = 2.0$, see **Supplementary Information, Note S1**), which remains a closed-loop strategy for precision navigation in small-scale systems **[19, 47-49]**. The comparison spans five geometries of increasing complexity, as shown in **Fig.1e**. These geometries test straight-line transport, discrete directional change, and sustained steering through continuous curvature. We evaluated MBRL policy over 20 episodes and PID over 10 episodes per geometry.

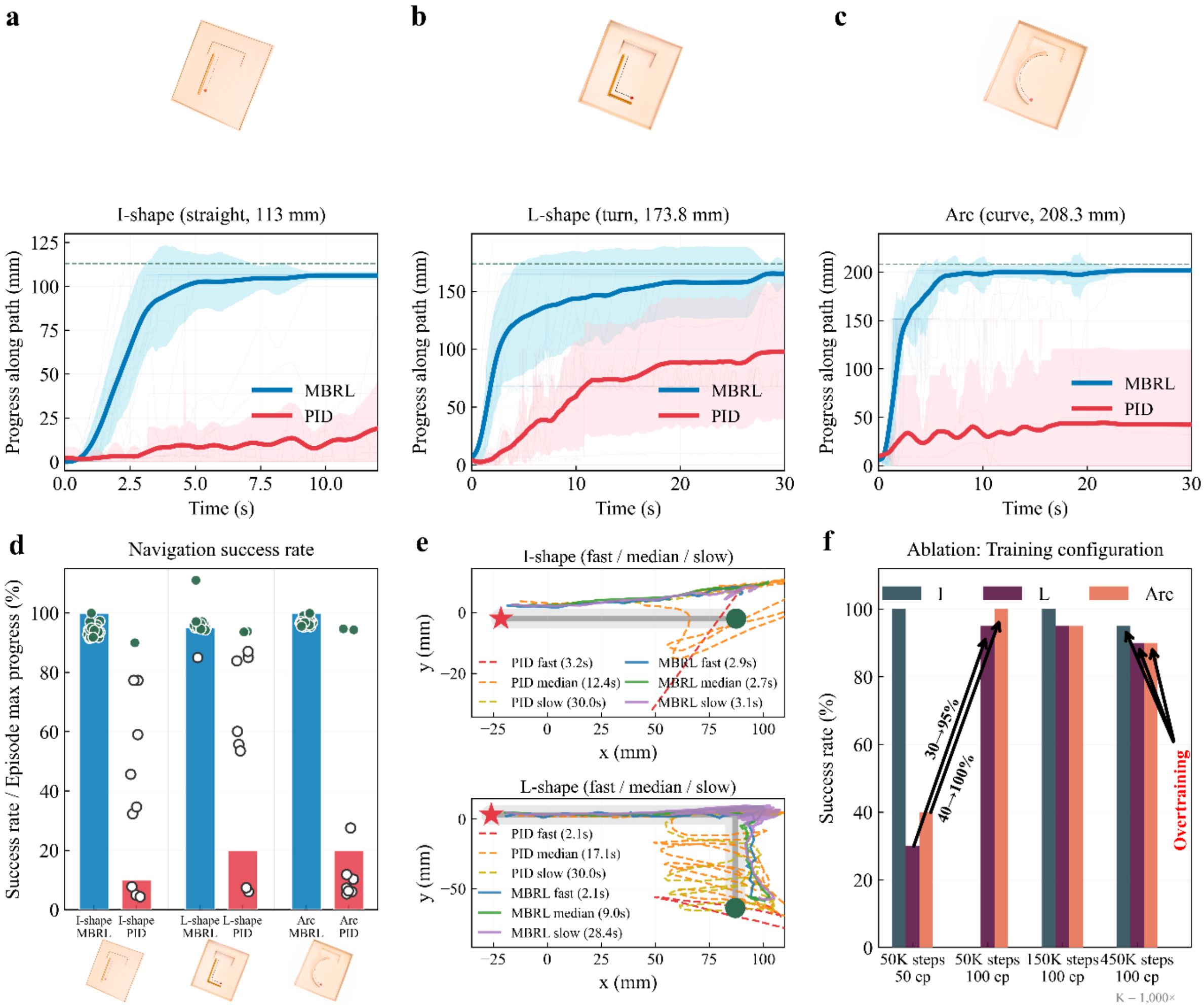

**Fig. 3 | Autonomous droplet navigation performance across three path geometries of increasing complexity**. **a–c,** Path progress as a function of time for the I-shape, L-in-shape and Arc-in-shape. Top: photographs of each fabricated path with the droplet (red) at the start position. Bottom: thick lines represent mean progress across episodes (MBRL: blue; PID: red); shaded regions indicate standard deviation. Thin lines show individual episode trajectories. The

dashed green line marks the goal. **d,** Navigation success rate (bar height) with individual episode maximum progress overlaid (filled green circles: success; open circles: failure). MBRL achieves 100% (I-shape), 95% (L-in-shape), and 100% (Arc-in) success rates compared to 10%, 20%, and 20% for PID across the respective geometries. **e,** Spatial (x, y) trajectories for three representative episodes per controller (fast, median, slow by completion time) on the I-shape (top) and L-in-shape (bottom) geometries. The gray band indicates the reference path. Green circle represents start, whereas red star shows goal. MBRL trajectories (blue shades) tightly follow the reference path, whereas PID trajectories (red shades) exhibit large lateral oscillations on the I-shape and chaotic looping at the L-in-shape corner. Episode durations are annotated for each trajectory. **f,** Success rate as a function of training gradient steps and number of training episodes for the L-shape geometry. Increasing training data from 50 to 100 episodes at fixed 50,000 gradient steps improves success rate from 40-100% for arc shape and 30-95% for L-shape. However, extended training to 450,000 steps on the same 100 episodes degrades performance, indicating overtraining in which the world model memorizes rather than generalizes from the training distribution.

MBRL achieves rapid, consistent navigation for I-shape (**Fig. 3a),** All 20 episodes reach the goal (**Movie S2**). The fastest episode completes in just 2.0 seconds. The mean time to reach the target is 3.5 seconds. The mean PID curve rises slowly and stabilizes around 15–20 mm, well below the goal. The individual PID trajectories (thin red lines) are scattered widely. PID episodes show only 10% success rate (**Movie S3**), and that single successful episode took 12.4 seconds. The mean PID progress is only 49.0 $\pm$ 34.3 mm, compared to MBRL's 106.3 $\pm$ 2.4 mm. The high standard deviation (34.3 mm, or 70% of the mean) reflects the PID controller's fundamental unreliability for this task. **Fig. 3b** shows L-in-shape progress, significantly difficult navigation task because the droplet must change direction at the corner while maintaining path adherence. MBRL navigates it in a mean of 7.7 s and as fast as 2.1 s (**Movie S4**) with 95% success rate. PID instead saturates at 75-100 mm, significantly at the corner region (**Movie S5**), with mean progress of 108.8 ± 54.6 mm, with 20% success rate. Most episodes are either stuck at the corner or overshoot off path, because the corner imposes a threshold for motion initiation and droplet velocity responds nonlinearly to tilt.

The Arc-in-shape (**Fig. 3c**) demands sustained steering rather than a single discrete turn. A 150° curve over 208.3 mm requires a policy that continuously adjusts the tilt vector to track a smoothly curving reference. The learned policy navigates it reliably with 100% success rate **(Movie S6)**, whereas PID progress saturates well below the goal **(Movie S7)**, with only 20% success. Its failure here is mechanistically distinct, rather than a single catastrophic overshoot at a corner, the PID accumulates tracking error continuously along the curve because its reactive corrections cannot anticipate the evolving tilt direction required by the changing path curvature. MBRL policy predicts the droplet response to tilt sequences over multiple steps, enabling anticipatory steering. The navigation success rates across all three geometries are summarized in **Fig. 3d**. The PID failures are distributed across the full range from 5% to 90%, reflecting irreproducible episodes and primarily unpredictable behavior.

To evaluate the generality of the learned strategy beyond wall-confined geometry, we tested the MBRL policy on the exterior of the L-shaped boundary (L-out, 287.2 mm), where the droplet navigates the corner without any physical confinement (**Movie S8**). In the inside configuration the walls serve as dual function: constrain lateral motion and provide a contact surface against which the agent's oscillatory depinning strategy (discussed below) generates the force reversals that overcome capillary adhesion at the corner. Neither mechanism survives in the outside configuration. The droplet approaches the corner in open space, and any overshoot carries it off the path with no wall to redirect it. The control problem therefore shifts from depinning to anticipatory deceleration and steering, requiring the policy to anticipate a high-speed approach to an unconfined turn and reduce tilt amplitude before it reaches to the corner. The learned policy achieves 90% success on L-out (**Movie S8**) against no successful episode for PID (**Movie S9**), and it adapts its actuation to each geometry (**Table 1**). On the L-in-shape the sign change rate (the fraction of consecutive steps in which the motor command reverses polarity) is 42 $\pm$ 10% for Motor 2 (Y-axis tilt) that drives lateral motion toward and away from the corner wall. This axis needs the most aggressive reversals to overcome contact-line adhesion at the wall and corner. On L-out-shape the same rate falls to 19 $\pm$ 4%, less than half, as Motor 2 shifts from rapid depinning oscillations to sustained directional commands (**Movie S8**). Motor 1 changes far less, from 27 $\pm$ 6% to 21 $\pm$ 4%, consistent with its role in forward transport across both geometries. We also tested the MBRL policy on the Arc-out path (249 mm) combines both demands, unconfined cornering and sustained curvature. MBRL again reaches 90% success (**Movie S10**) whereas PID succeeds in 20% of trials (**Movie S11**). Together these results show that the learned policy captures enough of the droplet physics to generalize across qualitatively distinct constraints, from discrete unconfined corners to sustained curvature and the loss of wall-assisted depinning, with no change to the algorithm or the reward structure.

Moreover, **Fig. 3e** reveals spatial trajectories of droplet navigation. On the I-shape, MBRL-based episodes stay within a mean lateral deviation of 5.8 $\pm$ 2.3 mm, whereas PID excursions span roughly 49 mm, nearly fivefold wider, in a zigzag that follows from its reactive waypoint tracking as it overshoots laterally, corrects, and overshoots in the opposite direction. The contrast sharpens at the L-shape corner, where MBRL executes a clean turn while PID loops up to 89 mm off path and in several episodes crosses itself repeatedly, having lost directional control. The nonlinear droplet response at the corner converts each waypoint correction into a wide arc, and PID does not recover its heading once past the turn.

To characterize the relationship between training data, gradient steps, and policy performance, we conducted an ablation across the three geometries **(Fig. 3f)** which separates the contribution of data from that of optimization. With 50 random exploration episodes and 50,000 gradient steps the policy already solves the I-shape but not the other shapes, indicating that random actions cover straight-line dynamics adequately while sampling corner and curvature interactions poorly. A second batch of 50 episodes, collected under the explore policy (see training RL section), lifts both to 95%. Raising the step count to 150,000 on the same data leaves the I-shape and L-shape unchanged and slightly degrades the Arc-in, and 450,000 steps degrades all three, most severely the Arc, as the policy begins to memorize individual episode trajectories. A modest training size of 50,000–150,000 gradient steps are sufficient for reliable policy learning across these geometries, and that extended training without additional data is unnecessary and actively detrimental.

**Discovery of physically motivated oscillatory depinning control**

We show that the MBRL agent developed an oscillatory control strategy that was neither programmed nor rewarded explicitly. The trained policy achieved a 95% success rate, with all successful trajectories converging along the vertical segment and navigating the L-in-shape to reach the goal despite substantial variation in completion time, from 2.1 s (fastest) to 28.35 s (slowest) (**Fig. 4a**). Rather than proposing smooth, monotonic tilt control actions, as a classical controller would do, the trained policy rapidly alternates motor currents at every time step, effectively vibrating the board at a characteristic frequency **(Movie S4)**.

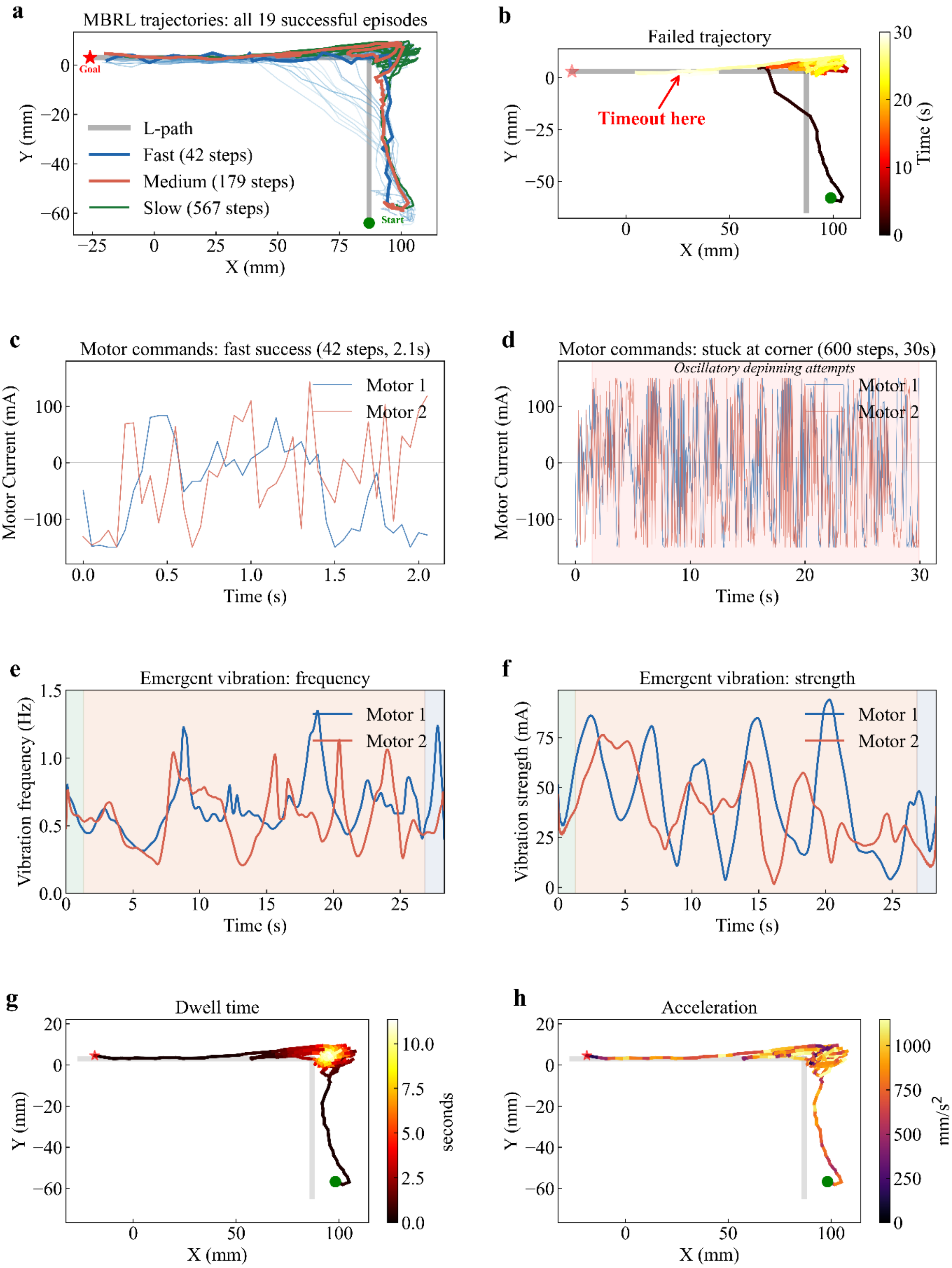

**Fig. 4 | Oscillatory control strategy learned by the MBRL agent. a,** Trajectories of all 19 successful episodes (light blue) overlaid on the L-shape path (gray), with three representative episodes highlighted: fast completion in 42 steps (2.1 s, dark blue), moderate completion in 179 steps (red), and slow completion in 567 steps (green). All trajectories converge along the

vertical segment and navigate the corner to reach the goal. **b,** Trajectory of the single failed episode (600 steps, 30 s), colored by elapsed time. The droplet traverses the vertical segment within the first 5 s (dark shading) but becomes trapped at the L-corner, where it oscillates in a confined region for the remaining 25 s (marked yellow) without reaching the goal. **c**, Motor current commands during the fastest successful episode (42 steps, 2.1 s). Both motors exhibit rapid sign reversals throughout the trajectory, alternating between approximately $\pm 150$ mA rather than maintaining steady unidirectional commands. **d,** Motor current commands during the failed episode (600 steps, 30 s). The shaded region marks the stagnation period at the corner; notably, the oscillatory command pattern persists throughout the stagnation, indicating active depinning attempts by the learned policy. **e,** Transient instantaneous vibration frequency of the motor current commands during the 567-step slow episode. Three phases of navigation are identified: start point to the corner (approach phase, green shading), stagnation at the corner (pink shading), and exit from the corner (escape phase, blue shading). Motor 1 (blue) exhibits repeated frequency excursions to 0.3–1.3 Hz during the corner phase, substantially above the 0.2–0.4 Hz baseline of Motor 2 (red). **f,** Instantaneous vibration strength (amplitude) of the motor commands over the same episode and phase decomposition. Motor 1 exhibits repeated amplitude surges exceeding 80 mA during the corner phase, consistent with escalating depinning attempts, while the approach and escape phases show comparatively suppressed amplitudes. **g,** Trajectory of the slow episode (567-steps) colored by local dwell time, defined as the cumulative duration the droplet spends within a 5 mm radius of each position. The bright cluster at the L-corner (dwell time exceeding 10 s) confirms prolonged stagnation at the geometric constriction. **h,** Same trajectory colored by instantaneous acceleration magnitude. Acceleration exceeding 1000 mm/s$^2$ concentrates at the corner region, coinciding spatially with the dwell time maximum, indicating that the agent applies its most aggressive oscillatory forcing precisely where the droplet encounters the greatest resistance.

To quantify the depinning oscillation strategy (**see Supplementary Information, Note S2 for details**), we decomposed the slow episode (28.35 s) into approach, corner and escape phases (**Fig. 4e, f, shaded regions**), based on the droplet's location. During the approach, Motor 1 oscillates at 0.4-0.8 Hz with vibration strength below 50 mA, gentle steering that clears the unobstructed vertical segment within 2 s, reflects the low transport resistance on unobstructed paths. This indicates that the policy learned that periodic force reversals assist transport even in the absence of geometric constrictions, consistent with the role of substrate vibration in reducing the pinning effect during droplet motion on any wetted surface. Oscillation therefore spans the whole path, and what distinguishes the corner is modulation, since the agent raises both frequency and amplitude when the droplet meets higher resistance and reduces them on unobstructed segments. In the corner phase, Motor 1 produces repeated depinning bursts in which instantaneous frequency rises from a 0.4-0.6 Hz baseline to 1.0-1.3 Hz (near t = 8, 18, 21 and 24 s) while vibration strength exceeds 75 mA and peaks at 85 mA. Motor 1 progressively assumes the dominant role, consistent with the horizontal forcing required to drive the droplet away from the corner wall, and both motors contribute with their relative dominance shifting according to the geometric demand. On exit, the droplet clears the remaining path within 2 s, mirroring the approach.

The spatial distribution of the droplet's response confirms the temporal analysis **(Fig. 4g, h)**. The droplet dwells for more than 10 s within a narrow region at the corner, against less than 2 s anywhere along the straight segments, and instantaneous acceleration peaks above 1000 mm/s$^2$ in that same region. Peak forcing therefore coincides with prolonged stagnation, which links the control escalation seen in the motor commands (**Fig. 4e, f**) to the physical response of the droplet. The policy is acting in closed loop, detecting stagnation from position feedback and handling with more intense oscillation, as discussed in the next section.

The oscillation analysis connects to the vibration-induced transport framework [**7, 8, 50**], in which externally applied sinusoidal substrate vibration directed droplet motion on chemical gradient surfaces where static forces could not overcome contact angle hysteresis. The frequency elevation and amplitude burst at the corner (**Fig. 4e, f**) are the learned equivalent of that prescribed vibration, exploiting periodic force reversals to break adhesive barriers that would arrest the droplet under steady unidirectional forcing. Our system advances beyond the original vibration-induced transport paradigm in two aspects. First, earlier studies **[7-9, 50-52]** vibrated the entire substrate uniformly at an operator-set frequency and amplitude. The agent instead decides when, where and how intense to oscillate, optimizing all three against the navigation objective, so forcing escalates at the corner and withdraws on the straight segments rather than being applied indiscriminately. Second, this position-dependent mobilization emerged from random exploration data alone, with no encoding of contact angle hysteresis, interfacial adhesion or the vibration mechanism in the reward, the observation space or the training objective. To our knowledge, this is the first MBRL agent to discover and spatially optimize a vibration-based mobilization strategy for liquid droplet manipulation.

In contrast, the single failed episode (**Fig. 4b, d**) shows the oscillation is intentional rather than incidental. Through 25 s of stagnation at the corner, the policy sustains oscillatory commands instead of a fixed output, consistent with persistent depinning attempts. The failure reflects insufficient oscillation amplitude to overcome the wall adhesion at the corner within the allotted 600 steps (30 s), not an absence of the mobilization strategy. The learned policy has therefore internalized the empirical relationship between oscillatory forcing and mobilization, and the policy deploys that strategy whenever its learned dynamics predict stagnation.

### Comparison of closed-loop and open-loop

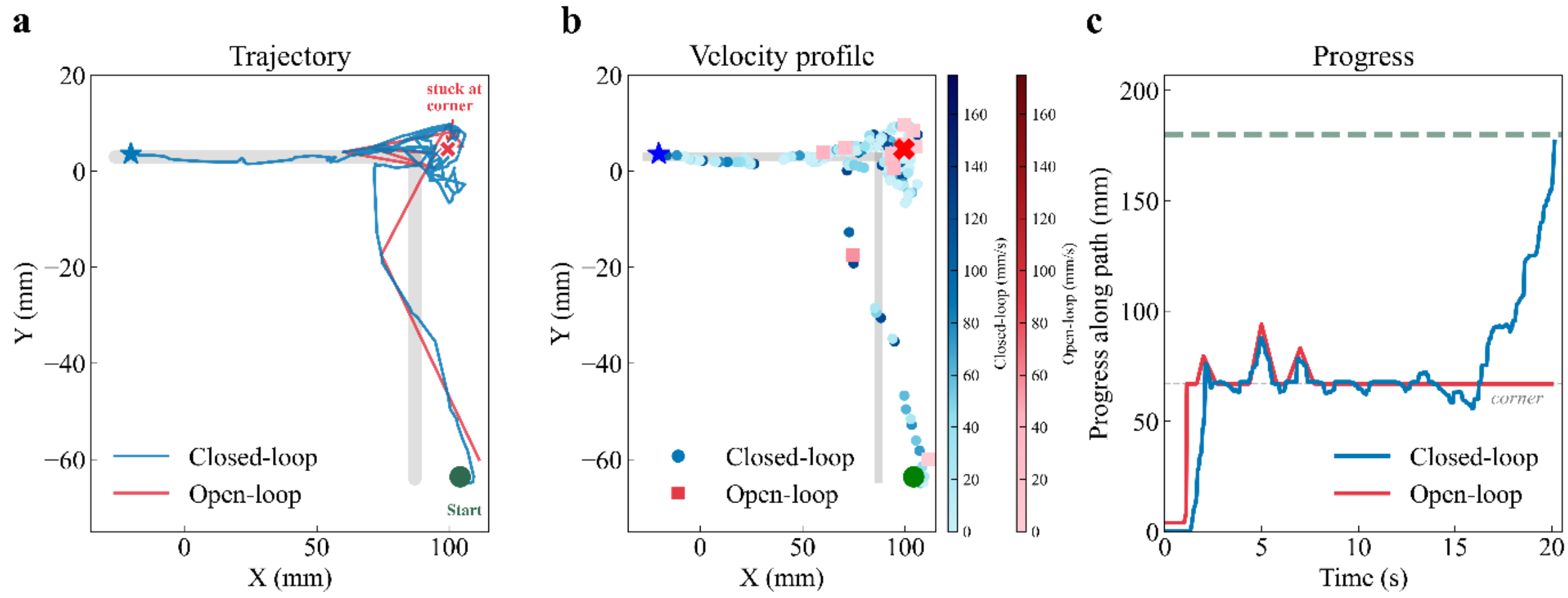

**Fig. 5 | Execution of closed-loop vs. open-loop of the learned MBRL policy on L-in-shape**. Motor commands recorded from a successful closed-loop episode (404 steps, 20.2 s) were replayed open-loop without camera feedback. **a**, Spatial trajectories: the closed-loop agent (blue) navigates the full L-shape path including the corner, while the open-loop replay (red) diverges at the corner. **b**, Velocity profiles of both executions overlaid on the L-in-path; closed-loop points (circles, blue gradient) and open-loop points (squares, red gradient) are colored by instantaneous speed. The closed-loop trajectory shows distributed high-velocity motion throughout the path, whereas open-loop motion concentrates near the corner before stopping.

**c,** Progress along the waypoint path. The closed-loop agent (blue) advances past the corner and reaches the target, while the open-loop replay (red) stops at the corner.

To verify that the learned policy relies on real-time feedback rather than memorized action sequences **(Fig. 5)**, we replayed the motor commands from a successful episode without camera feedback (**Movie S12**). The droplet trajectory diverges from the planned path within 3 seconds of execution and fails to negotiate the L-shape corner. Along the initial straight segment, the open-loop trajectory tracks the closed-loop path closely, because the dynamics there are nearly linear and prediction errors remain small. However, at the corner, where the required control direction changes abruptly and corner pinning introduces discontinuous resistance, accumulated position errors become critical. The closed-loop agent observes its actual state every 50 ms and corrects for these deviations, whereas the replay applies commands computed for positions the droplet no longer occupies. The learned policy is therefore accurate enough for single-step evaluation within a feedback loop but not for multi-step open-loop planning. Microfluidic systems where stochastic surface interactions, evaporation, and contact-angle pinning introduce irreducible model uncertainty that can only be mitigated through continuous state estimation and feedback.

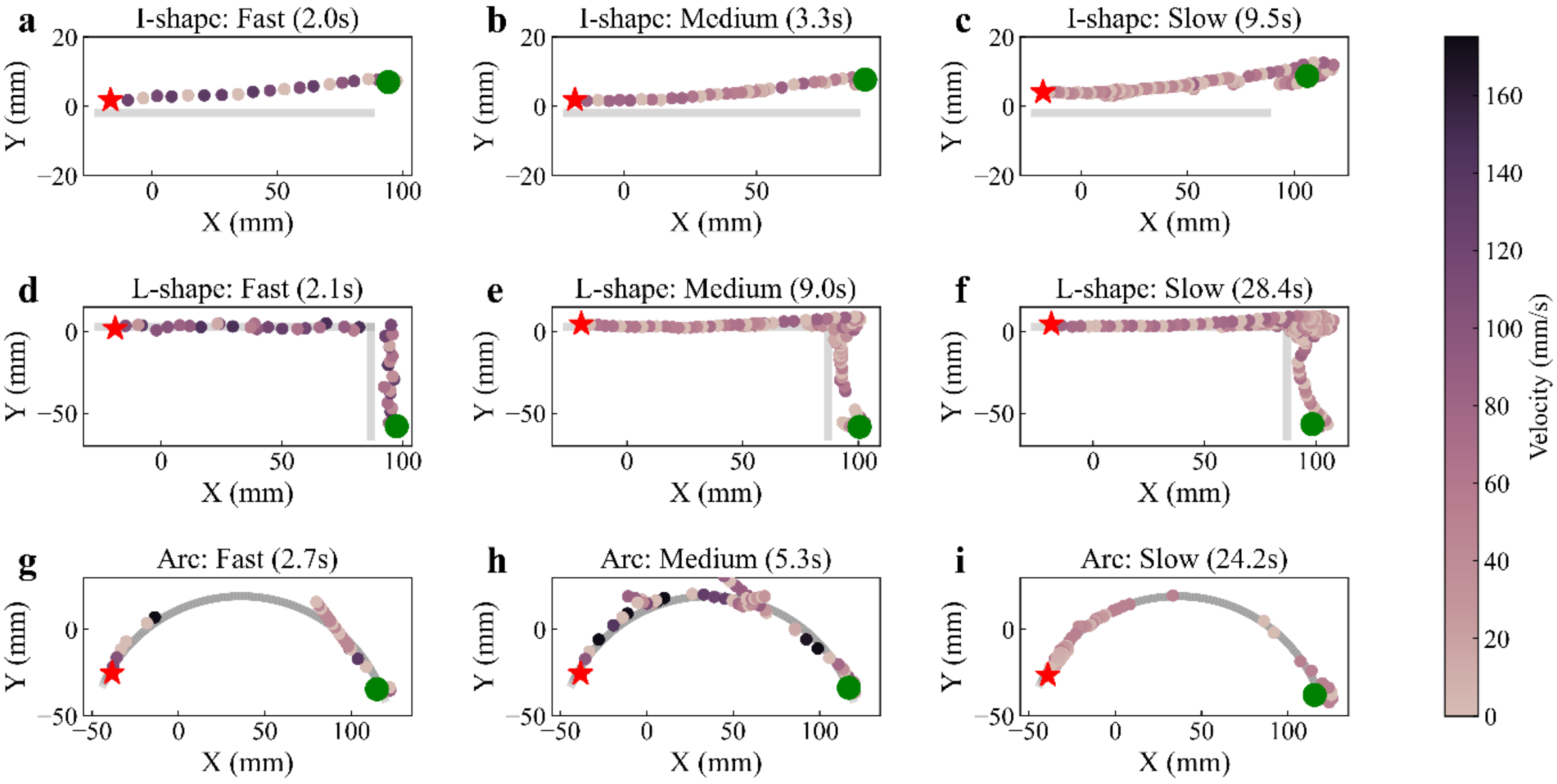

**Fig. 6 | Velocity-mapped trajectories across three geometries.** Columns correspond to fast, medium, and slow episodes for each geometry. **a–c,** I-shape: the droplet maintains near-uniform velocity along the straight path. **d–f,** L-in-shape: velocity is highest along the straight segments and drops sharply at the corner, where capillary pinning arrests the droplet. **g–i,** Arc-in-shape: velocity decreases progressively along the curve as the droplet must continuously negotiate the changing wall geometry. Green circle: start position, and red star: goal.

## Velocity profile

The velocity heatmap (**Fig. 6**) demonstrates that the agent's navigation strategy adapts to the local geometric constraints of each path. Uniform velocity on the I-shape indicates that straight-segment motion is viscous drag-limited rather than geometric barriers, whereas the L-in-shape produces localized arrest at the corner before eventual depinning. On the Arc-in-shape, the continuous curvature imposes distributed transport resistance along the entire path, resulting in a progressive velocity reduction. This contrast between localized arrest and progressive deceleration shows that the policy modulates its output to the spatially varying resistance of each geometry rather than applying one fixed strategy.

### Zero-shot transfer learning to unseen geometries

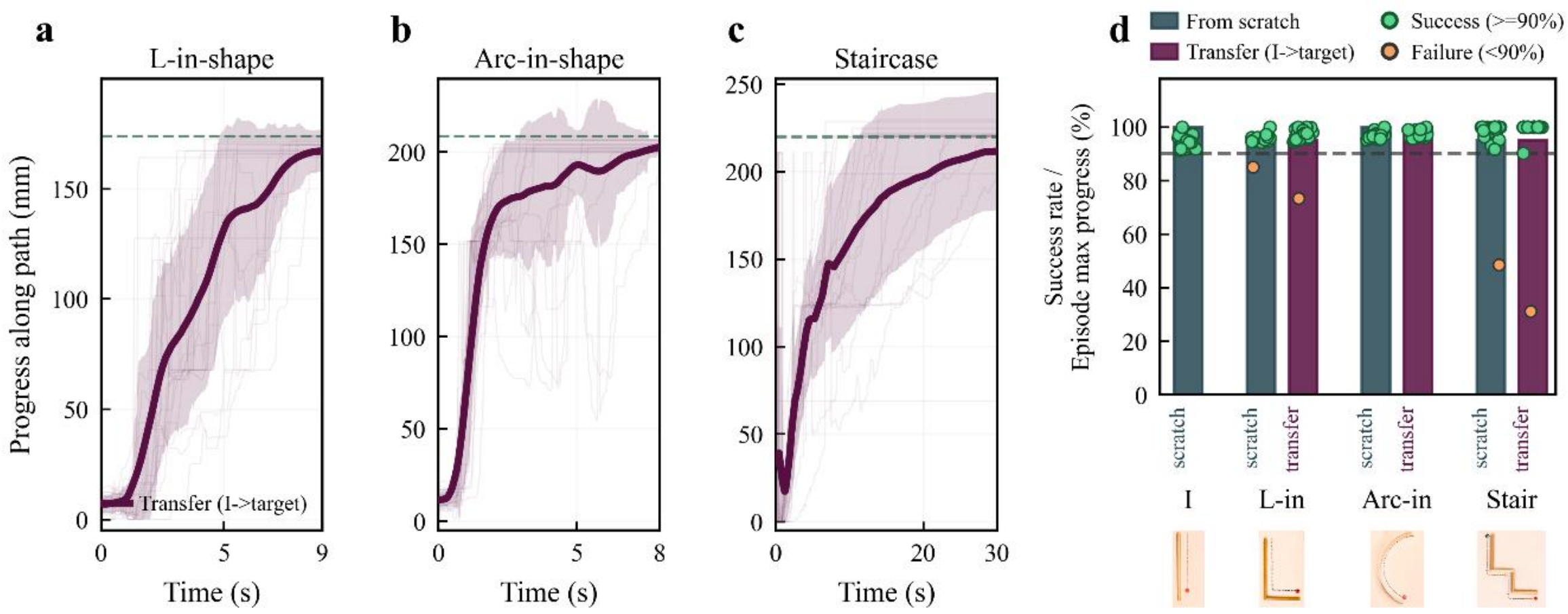

**Fig. 7 | Zero-shot transfer of the I-shape policy to unseen geometries. a-c,** Progress along the path against time for the transferred policy on the L-in (**a**), Arc-in after fine-tuning on 20 additional episodes (**b**) and Staircase (**c**) geometries. Thick lines show the mean across 20 deployment episodes, shaded regions the standard deviation, thin lines individual episodes, and the dashed green line the goal. Corresponding from-scratch curves are shown in **Fig. 3a-c**. **d,** Success rate for from-scratch training against zero-shot transfer across all three target paths, with individual episode maximum progress overlaid (green, success; orange, failure) and the dashed line marking the 90% threshold. Path outlines below indicate each geometry.

To determine whether the learned policy captures the underlying droplet dynamics rather than overfitting to a single geometry, we evaluate its ability to transfer across unseen paths. We deploy the policy trained only on the I-shape directly on three previously unseen geometries, namely the L-in, the Arc-in, and a Staircase path. **Fig. 7** compares zero-shot transfer against policy trained from scratch on each target path, with 20 deployment episodes per condition.

The transferred I-shape policy generalizes remarkably well. On the L-in-shape it achieves 95% success (**Movie S13**), matching the from-scratch policy exactly, even though the policy never saw a corner during training (**Fig. 7a**). This is a strong demonstration from a control standpoint, because the corner requires the droplet to reverse its lateral tilt and the controller to reject an overshoot that a classical law would accumulate. The transferred policy handles the turn cleanly without carrying the droplet off the path. The Arc-in-shape is the more interesting case. When the I-shape policy is transferred directly, it traverses roughly 85% of the arc before failing, because sustained curvature demands a continuously evolving tilt vector that straight-line training never exposed. We therefore collect 20 additional episodes on the Arc-in-shape and train the existing policy, which learns the missing curvature-tracking behavior and achieves 100% success (**Movie S14**), the same as the from-scratch baseline (**Fig. 7b**). Even where zero-shot transfer is incomplete, the transferred model provides a strong initialization that converges on a fifth of the data required to train anew. The Staircase combines several discrete turns in sequence with outward portions where no wall redirects an overshoot and the margin for error is small. Here the policy decelerates and steers preemptively, and again reaches 95% success (**Movie S15**), which indicates that it encodes the droplet response to tilt sequences and not a task-specific waypoint schedule (**Fig. 7c**). Notably, the Staircase can be viewed as a repetition of a single unit cell, and the I-shape policy generalizes across every cell without any decay in performance. This is characteristic of an infinite-horizon solution, since the same stationary policy remains effective at each repeated segment regardless of number of turns preceding it. The policy therefore scales to arbitrarily long staircase paths without retraining, because its control law is time-invariant and depends only on the local droplet state, not on absolute position along the path.

The learned policy decouples the cost of learning from the number of target tasks. The soft, deformable nature of the droplet makes each task highly nonlinear and difficult to model analytically, so a controller that must be redesigned for every path would be impractical. Transfer learning removes this redesign entirely. A single policy trained on the simplest geometry serves as a reusable dynamics prior that either deploys directly or adapts to a new path on 20 episodes.

**Discussion**

The use of RL enables autonomous navigation of a liquid droplet on an oil-lubricated surface through complex channel geometries, a problem that has not been addressed by prior control strategies. A PID controller operating on the same hardware achieves no more than 20% success across these geometries, demonstrating that classical linear feedback is inadequate for this nonlinear and partially observable system. The comparison across other systems, spanning magnetic **[20]**, acoustic **[22],** and gravitational actuation of solid objects **[36]**, these works indicate that RL is emerging as a general-purpose control paradigm for autonomous micro-scale navigation. A central distinction between our approach and prior RL microrobotics work is the absence of any simulation environment. Modelling the droplet-on-oil system would require resolving the Navier-Stokes equations coupled with dynamic contact-line models, oil film depletion kinetics and meniscus deformation at geometric boundaries, which would be expensive and still unlikely to capture the behavior of the real system. Model predictive control inherits this difficulty, since its performance is bounded by the fidelity of the model it optimizes over, and optimizing over a model of this complexity is not feasible within the 50 ms control interval. Instead, we train the policy on physical episodes collected under random and explore-policy actions and obtain convergent policies from 50 to 100 episodes per geometry. Beyond the navigation results, the agent discovered an oscillatory depinning strategy with no encoding of interfacial physics in the reward, the observation space or the training objective. The open-loop ablation is equally informative. When the exact motor command sequence from a successful closed-loop episode is replayed without camera feedback, the droplet fails to navigate. Learned controllers for microfluidic systems must therefore operate in closed loop, because stochastic surface interactions, oil film depletion and pinning introduce irreducible model uncertainty that open-loop execution cannot accommodate. Zero-shot transfer completes the autonomous droplet navigation strategies, with a straight-line policy generalizing to unseen turns and curves without retraining from scratch.

Looking ahead, several directions extend from the current results. We will extend single-droplet navigation to multi-droplet systems, placing droplets on the same board with distinct target destinations and shapes, where agents learn to coordinate the timing and direction of tilt commands so that each droplet progresses toward its own goal. This would be an underactuated multi-agent control problem where the action space remains fixed, while the state space and coordination complexity grows. Finally, the framework presented here is not limited to gravitational actuation on oil-lubricated surfaces. The core components, an overhead camera for state estimation, and a DreamerV3 world model for dynamics learning, are modality-agnostic. The same approach could be adapted to electrowetting-based droplet routing by replacing the current action space with electrode voltage patterns. More broadly, any micro-scale system in which the actuation-to-response mapping is nonlinear, partially observable, and expensive to simulate from first principles is a candidate for MBRL-based autonomous control. The demonstration that a liquid droplet, the simplest and most ubiquitous soft-matter object, can be navigated autonomously through complex geometries using only visual feedback and MBRL has substantial untapped potential in microfluidics and chemical engineering.

## Methods

### Experimental platform

Our in-house DropletRunner platform features a commercial wooden Labyrinth maze (BRIO Labyrinth, purchased from BRIO) as the tilting substrate for droplet navigation. Two Dynamixel XL330-M077-T servo motors (purchased from ROBOTIS) operating in current control mode actuate the board about orthogonal axes via the Labyrinth's built-in mechanism. Motor 1 controls tilt along the x-axis of the estimator coordinate frame and Motor 2 along the y-axis; both accept signed 16-bit current commands in the range $[-150, +150]$. The motors are interfaced to the host computer through a U2D2 USB-to-Dynamixel adapter (ROBOTIS) mounted on a U2D2 Power Hub Board Set (ROBOTIS), which provides regulated power distribution. Three-pin Robot Cable X3P connectors (ROBOTIS, 10-pack) link each motor to the hub. The entire assembly is powered by a 5V3A USB power supply. (see **Fig. S1**)

**Navigation surface fabrication**. We fabricated custom flat surfaces to replace Labyrinth's original perforated board, allowing droplet transport to be evaluated across different paths. The base board is a flat PLA plate printed to match the Labyrinth frame dimensions (**Supplementary Information, Note S3**).

**I-, L-, and Arc-shape geometries.** For the straight-channel (I-shape) and right-angle (L-shape) paths, we bonded thin plywood strips onto the PLA base using adhesive to form raised wall boundaries that confine the droplet to the intended corridor. The channel centerline was marked on the board with dotted black lines using a fine-tip permanent marker to provide a visual reference for waypoint alignment during calibration. The I-shape channel length is 113 mm, and the L-shape channel length is 173.8 mm. We designed the 150° arc channel (radius of curvature 80 mm) in SolidWorks, exported the geometry as an STL file and 3D-printed the curved wall segments in PLA. The printed arc pieces were bonded to the flat base board with adhesive, producing a continuous curved corridor.

**Surface preparation.** After assembly, the entire navigation surface was coated with a thin layer of AR 20 silicone oil (viscosity 20 mPa-s; 500 ml bottle, purchased from Sigma-Aldrich). The oil was dispensed from a pipette and spread uniformly, producing a lubricating film. This oil layer serves two functions: it reduces rolling friction between the water droplet and the PLA substrate, enabling rapid sliding dynamics, and it creates a low-hysteresis interface that permits the droplet to respond to small tilts. Over extended experimental sessions (>48 hours), localized oil depletion can occur, producing regions of increased pinning; the board is re-oiled between experimental batches to restore consistent surface conditions.

**Waypoint selection and path definition.** Each geometry is described by an ordered sequence of waypoints $W = w_0, w_1, \ldots, w_N$, where each waypoint $w_j = (x_j, y_j)$ is expressed in the OCamCalib-corrected estimator coordinate frame (units: mm). Waypoints were selected manually by placing the droplet at salient positions along the intended path and stored as a JSON file. The path defined by the waypoint sequence also serves as the reference for episode termination. At each control time step, the perpendicular distance from the droplet to the nearest line segment connecting consecutive waypoints is computed. If this deviation exceeds a geometry-specific threshold $d_{max}$ (typically 30 mm), the episode is terminated as a failure. An episode succeeds when the droplet reaches within $d_{goal}$ of the final waypoint $w_N$, and times out if the maximum number of steps $T_{max}$ is exceeded without satisfying either condition. These criteria are applied identically to both the MBRL agent and the PID baseline.

**Algorithm 1.** MBRL offline training and autonomous navigation

**Input:** Waypoint sequence $W = \{w_0, w_1, \ldots, w_N\}$, camera feed, episodes per stage $E_{\text{stage}}$, gradient steps per cycle $G$, maximum motor current $A_{\max} = 150$, success threshold $\tau$.

**Shared subroutine 1: SELECTACTION(mode, $I_t, v_t, r_t$)**

1: **if** `mode` = `random` **then**
2: $a_t^{\text{raw}} \sim \mathcal{N}(0, \sigma^2)$ ▷ *random motor command*
3: $a_t \leftarrow 0.7 a_{t-1} + 0.3 a_t^{\text{raw}}$ ▷ *temporal smoothing*
4: $a_t \leftarrow \text{CLAMP}(a_t, -A_{\max}, +A_{\max})$ ▷ *actuator limits* $[-150, +150]$
5: **else if** `mode` = `eval` **then**
6: $a_t \leftarrow A_{\max}\, \pi_\theta([I_t, v_t], r_t)$ ▷ *policy inference, then scale to motor currents*
7: **else if** `mode` = `explore` **then**
8: $a_t \leftarrow A_{\max}\, \pi_\theta([I_t, v_t], r_t;$ `mode=explore`$)$ ▷ *policy and exploration noise*
9: **end if**
10: **return** $a_t$

**Shared subroutine 2: RUNEPISODES($E$, mode, saveEpisodes)**

11: **for** episode $= 1$ to $E$ **do**
12: Place droplet at the start position manually
13: **for** $t = 0$ to $t_{\max}$ **do**
14: $H \leftarrow \text{GETCAMERAFRAME}()$ ▷ *RGB frame*
15: $\mathbf{p}, \alpha_{\text{tilt}}, \beta_{\text{tilt}} \leftarrow \text{STATEESTIMATOR}(H)$ ▷ *position $(x, y)$ and board tilt angles*
16: $I_t \leftarrow \text{RESIZEANDCROP}(H, 64 \times 64 \times 3)$ ▷ *image observation*
17: $r_t, \text{done}, \ell_{1:5} \leftarrow \text{PATHTRACKER.UPDATE}(\mathbf{p})$ ▷ *reward, termination, 5 lookahead points*
18: $v_t \leftarrow [p_x, p_y, \alpha_{\text{tilt}}, \beta_{\text{tilt}}, \ell_{1:5}]$ ▷ *14D state vector*
19: $a_t \leftarrow \text{SELECTACTION}(\text{mode}, I_t, v_t, r_t)$
20: **if** `saveEpisodes` **then**
21: Store $(I_t, v_t, a_t, r_t,$ `is_first`, `is_terminal`$)$ as transition
▷ *`is_first` is True if $t = 0$; `is_terminal` is True when the episode ends*
22: **end if**
23: Send motor commands $[C_1, C_2] \leftarrow [a_t^{(1)}, a_t^{(2)}]$
24: **if** goal reached **or** deviation $> d_{\max}$ **or** $t \geq t_{\max}$ **then**
25: **if** `saveEpisodes` **then**
26: Save episode as `.npz`
27: **end if**
28: **break**
29: **end if**
30: **end for**
31: **end for**

**Stage 1: Random exploration (CPU laptop, `data_collector`)**

32: RUNEPISODES($E_{\text{stage}}$, `random`, `True`)
33: **repeat**

**Stage 2: World-model training (GPU cluster, `train_offline`)**

34: Transfer all collected `.npz` episodes to GPU cluster via `scp`
35: Load episodes into replay buffer
36: Normalize stored actions: $a_t \leftarrow a_t / A_{\max}$ ▷ $[-150, +150] \rightarrow [-1, +1]$
37: **for** step $= 1$ to $G$ **do**
38: Sample a batch of 16 sequences $\times$ 64 timesteps
39: Update world model: encoder, GRU, decoder, reward head, continue head
40: Update actor $\pi_\theta$ to maximize imagined cumulative reward
41: Update critic $V_\psi$ using imagined $\lambda$-returns
42: **end for**
43: Save checkpoint containing world model, actor, and critic

**Stage 3: Evaluation and guided data collection (CPU laptop)**

44: Transfer checkpoint to CPU laptop
45: RUNEPISODES(20, `eval`, `False`) ▷ *no exploration; no episodes saved*
46: Compute success rate over 20 episodes
47: **if** success rate $< \tau$ **then**
48: RUNEPISODES($E_{\text{stage}}$, `explore`, `True`) ▷ *policy-guided exploration episodes are saved*
49: **end if**
50: **until** success rate $\geq \tau$
51: **Policy converged**; continue using `policy_runner` for autonomous deployment

**Algorithm 2.** PID waypoint-tracking controller

**Input:** Waypoint sequence $W = \{w_0, w_1, \ldots, w_N\}$, PID gains $K_p$, $K_i$, $K_d$, sign parameters $\sigma_1 = +1$ and $\sigma_2 = -1$, thresholds $d_{\text{thresh}}$, $d_{\text{goal}}$, $d_{\max}$, and $T_{\max}$.

```
 1: Initialize j ← 0,  e_prev ← (0,0),  I ← (0,0),  t ← 0
 2: repeat  at each control step, Δt = 0.05 s
 3:     p ← read droplet position from estimator          ▷ current position (x, y)
 4:     while ‖p − w_j‖ < d_thresh and j < N do
 5:         j ← j + 1                                      ▷ advance waypoint
 6:     end while
 7:     e ← w_j − p                                        ▷ positional error
 8:     I ← CLAMP(I + e Δt, −I_max, +I_max)                ▷ anti-windup
 9:     ė ← (e − e_prev)/Δt                                ▷ derivative error
10:     e_prev ← e
11:     u_1 ← σ_1(K_p e_x + K_i I_x + K_d ė_x)             ▷ Motor 1
12:     u_2 ← σ_2(K_p e_y + K_i I_y + K_d ė_y)             ▷ Motor 2
13:     u_1 ← CLAMP(u_1, −150, +150)
14:     u_2 ← CLAMP(u_2, −150, +150)
15:     Send motor commands [u_1, u_2]
16:     t ← t + 1
17:     if ‖p − w_N‖ < d_goal then
18:         return SUCCESS
19:     else if dist(p, path) > d_max then
20:         return FAILURE due to deviation
21:     else if t ≥ T_max then
22:         return FAILURE due to timeout
23:     end if
24: until episode terminates
```

**Data availability**

No public or custom datasets were used for the current study. All data required to reproduce the results reported in this work are provided within the main article, the Supplementary Information, and the accompanying GitHub repository. Data associated with this study is available from the corresponding author, MVK, upon reasonable request. All experimental movie links (Movie S1-S15) are provided in supplementary information (SI).

**Code availability**

All the data processing and analysis were performed using custom Python codes. The source code used for the postprocessing (analysis) and droplet navigation control in the physical experiments is available via GitHub at https://github.com/rajneeshanand/DropletRunner.

**Acknowledgements**

We acknowledge Lehigh University's High-Performance Computing (HPC) resources for training. We also acknowledge the assistance provided by Simeon Krizan (Lehigh University Design Lab manager) and Farheen Akhtar (Lehigh University, PhD student) in 3D printing and modeling the CAD designs. We thank Fatima Tanveer (Lehigh University STEM summer intern) for video editing and assistance with Arc-out-shape experiments.

**Author contributions**

RA and MVK conceived the research. RA designed experiments, methodology, and fabricated the board. RA wrote the code, integrated the physical setup, and performed all the experiments. RA and MVK worked on the analysis of data. RA and MVK wrote and reviewed the manuscript. MVK supervised the research and secured the funding. Both authors discussed the results and commented on the manuscript.

**Competing interests**

The authors declare no competing interests.

# Autonomous Droplet Navigation via Model-Based Reinforcement Learning

Rajneesh Anand[a] and Mayuresh V. Kothare[a,*]
[a]*Department of Chemical and Biomolecular Engineering, Lehigh University, Bethlehem, PA 18015, USA.*
*Corresponding author. E-mail: mvk2@lehigh.edu

## Supplementary Information (SI)

**This document includes:**

**I. Supplementary notes S1 to S3**

- **Supplementary Note S1**: System overview describes the four main components: physical setup, software architecture, control unit, training procedure, and PID workflow.
- **Supplementary Note S2**: Oscillation analysis. It describes the methodology used to extract instantaneous frequency and amplitude as a function of time from the motor-current signals.
- **Supplementary Note S3**: Fabrication of the 3D-printed PLA board. It describes the complete fabrication procedure.

**II. Supplementary figures S1 to S6**

- **Supplementary Figure S1**: DropletRunner Hardware assembly for autonomous droplet navigation.
- **Supplementary Figure S2**: State estimation from overhead camera.
- **Supplementary Figure S3**: GPU training benchmark for offline training.
- **Supplementary Figure S4**: Raw motor-current signals, full IMF decomposition, and Motor 2 Wu–Huang test.
- **Supplementary Figure S5**: EMD decomposition and three-stage IMF selection for Motor 1.
- **Supplementary Figure S6**: Fabrication workflow for navigation surface boards.

**III. Supplementary tables S1 and S2**

- **Supplementary Table S1:** Hardware components and their use
- **Supplementary Table S2:** Architecture, hyperparameters, inputs, and outputs of the DreamerV3 MBRL agent

**IV. Legends for supplementary movies S1 to S15**

- **Movie S1**: Manual droplet navigation.
- **Movie S2**: MBRL autonomous navigation, I-shape.
- **Movie S3**: PID autonomous navigation, I-shape.
- **Movie S4**: MBRL autonomous navigation, L-in-shape.
- **Movie S5**: PID autonomous navigation, L-in-shape.
- **Movie S6**: MBRL autonomous navigation, Arc-in-shape.
- **Movie S7**: PID autonomous navigation, Arc-in-shape.
- **Movie S8**: MBRL autonomous navigation, L-outside-shape.
- **Movie S9**: PID autonomous navigation, L-outside-shape.
- **Movie S10**: MBRL autonomous navigation, Arc-outside-shape.
- **Movie S11**: PID autonomous navigation, Arc-outside-shape.
- **Movie S12**: Closed-loop versus open-loop, L-in-shape.
- **Movie S13**: Zero-shot transfer from I-shape to L-in-shape.
- **Movie S14**: Zero-shot transfer from I-shape to Arc-in-shape.
- **Movie S15**: Zero-shot transfer from I-shape to Staircase.

**V. SI References**

## Supplementary Note S1. System Overview

The experimental setup for the autonomous navigation of droplet consists of three main components: the physical (hardware) setup, software (code) architecture, and control unit.

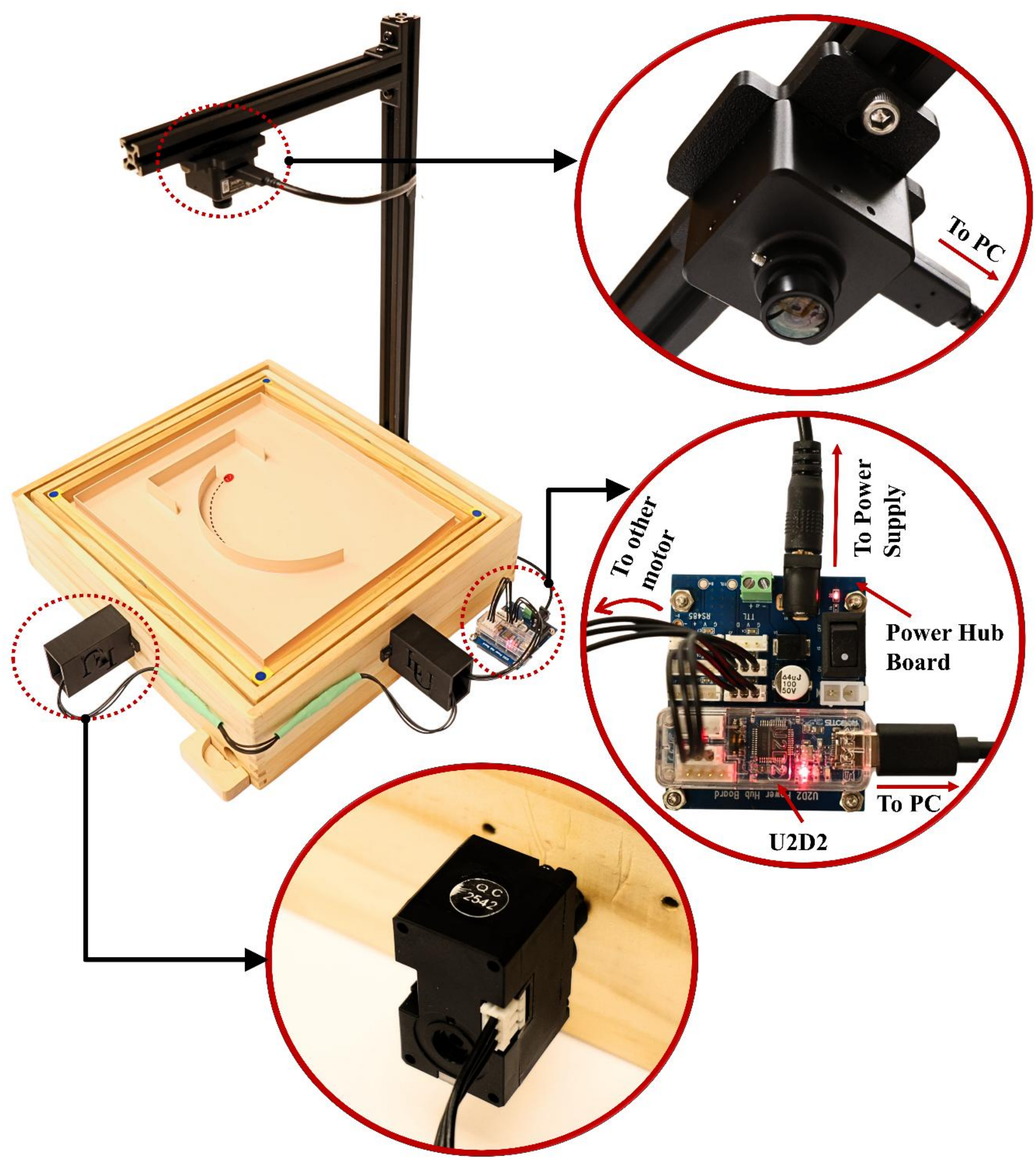

**Supplementary Figure S1 | DropletRunner Hardware assembly for autonomous droplet navigation.** The experimental platform comprises a 3D-printed PLA board mounted on a two-axis tilting platform actuated by two motors. An overhead camera (top right inset) provides visual feedback to the control PC via USB. The motors are daisy-chained through a U2D2 communication interface and powered via a Power Hub Board (middle right inset), which connects to a power supply and the control PC. A red-dyed water droplet on the oil-lubricated

board surface and four blue corner markers for homography-based position estimation are visible on the board. The bottom inset shows a close-up of the Dynamixel motor.

## 1. Physical Setup

DropletRunner is a physical model-based reinforcement learning (MBRL) system that autonomously navigates a liquid water droplet through a labyrinth maze. The droplet sits on an oil-coated surface and is actuated by tilting the board along two axes using Dynamixel servo motors. An overhead camera tracks the droplet position in real-time, and a DreamerV3 agent learns to control the tilt angles to guide the droplet from a start position to a target location. The system is inspired by and adapted from **[1]**, which used a similar MBRL framework to navigate a steel ball through a commercial labyrinth game. DropletRunner replaces the steel ball with a liquid droplet, introducing unique challenges: the droplet cannot fall into holes (unlike a ball), but it can break apart, is sensitive to surface tension effects, and moves more slowly due to viscous drag on the oil film.

Our in-house DropletRunner platform was built by modifying a commercial BRIO Labyrinth board game. The two manual control knobs were removed by prying them off with a screwdriver. Two Dynamixel motors were coupled to the exposed tilt shafts using couplers secured with provided screws. Pilot holes were drilled into the Labyrinth frame, and 3D-printed motor housings were slid over the motors and bolted to the frame to fix the motors in place. An overhead camera mount was constructed from T-slotted aluminum extrusion profiles with a 3D-printed mounting plate to hold the camera at a fixed height providing a full view of the board surface. Both motors were daisy-chained using a ROBOT X3P cable extended by splicing additional wire between the corresponding pins (soldered and covered with heat shrink, light green color). The motor chain was connected to a U2D2 USB-to-Dynamixel adapter mounted on a U2D2 Power Hub Board, which was powered by a 5V 3A USB supply. The U2D2 and camera were connected to the host laptop via USB, final closed-loop hardware chain shown in **Supplementary figure. S1** and component details are provided in **Supplementary Table S1.** (All CAD files are provided in GitHub.)

## 2. Software architecture

The system uses ROS2 (Robot Operating System 2) as the communication middleware, with four independent nodes running in separate terminals. Each node publishes or subscribes to ROS2 topics, enabling asynchronous, decoupled operation. The software is organized into three ROS2 packages: droplet_camera, droplet_motor, and droplet_state. The training side runs outside ROS2 entirely. A GitHub **[1]** fork of DreamerV3 is used as a library. The training script (train_offline.py) loads collected episode *.npz* (to store multiple NumPy arrays into a single file) files, converts them to DreamerV3's replay buffer format ({timestamp}-{uuid}-{successor}-{length}.npz), and runs gradient updates on the world model, actor, and critic networks. **Fig. 2** in the main paper demonstrates details for training.

**Motor Drivers.** The droplet_motor node deserves additional comments for their complexities in this study. This node controls the two Dynamixel XL330-M077 motors via the U2D2 USB-to-serial adapter. It operates in current control mode, where the command value directly sets the motor current (proportional to torque), rather than velocity or position mode. Dynamixel motors have three main control modes: position (go to angle X), velocity (spin at speed Y), and current (apply torque Z). In current control mode (mode 0), the command value directly sets the electrical current flowing through the motor coils. More current means more torque, which means stronger tilting force. The action space (a) is continuous and two-dimensional (**Eq. 1**).

$$a_t = [a_1,\ a_2] \in [-1,1]^2 \tag{1}$$

where $a_1$ and $a_2$ are linearly scaled to motor current commands $C_1$ and $C_2$ in the range [−150, +150] mA where $C_i = 150 \times a_i, i \in \{1,2\}$. Motor 1 ($C_1$) controls X-axis tilt and Motor 2 ($C_2$) controls Y-axis tilt. Together, they impose a two-dimensional gravitational body force on the droplet through differential inclination of the board surface. The relationship is nearly linear: current is proportional to torque. **[1]** uses the same Dynamixel motors (MX-12W) but in velocity control mode, where authors **[1]** command a rotational speed rather than a torque. We switched to current control because: (a) it gives more direct control over the tilt force, (b) there is no internal PID loop adding delay, and (c) it is simpler to implement (one register write). For a tilting board, torque control is more natural because we want to control how hard the board pushes the droplet, not how fast the motor spins. The motors are rebooted via the Dynamixel SDK reboot command every time the node starts. This clears any latched error states (overcurrent, overtemperature) that would otherwise prevent the motor from responding to commands. This is essential because after ~25-30 continuous episodes, Motor 1 tends to latch into an error state due to sustained current draw.

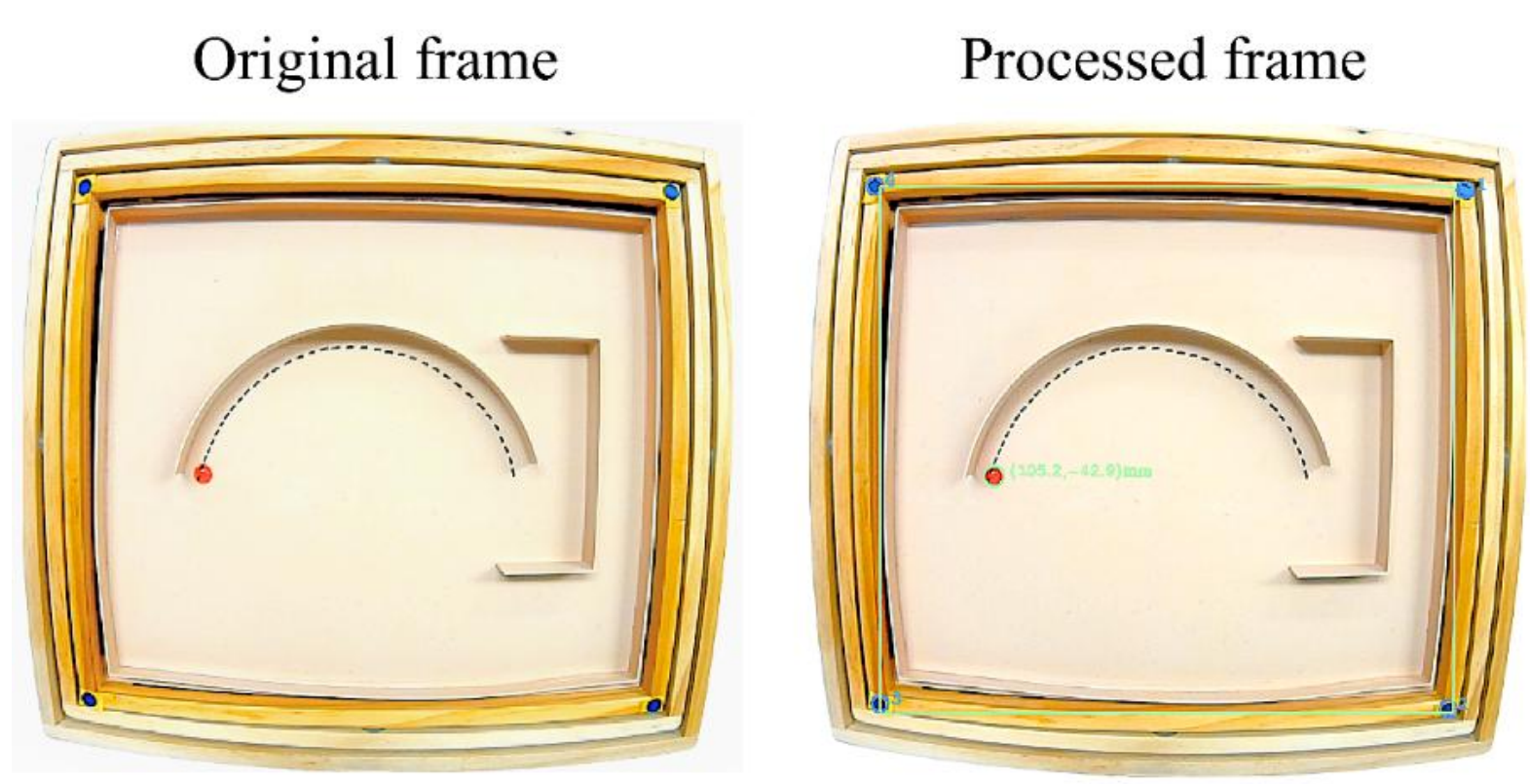

**Supplementary Figure S2 | State estimation from overhead camera**. Left: raw camera frame capturing the tilting board with an arc-shaped channel. The red-dyed droplet and four blue corner markers are visible. Right: processed frame with the estimated droplet position displayed in board-frame millimeters, computed from HSV color masking, OCamCalib distortion correction, and corner-based homography.

**State Estimation.** The droplet position is estimated from an overhead camera through a six-stage pipeline implemented as a ROS2 node. (i) Each camera frame is converted to HSV color space, and a tuned color mask isolates the red-dyed water droplet within a predefined region of interest. The droplet centroid is extracted via a color-segmentation pipeline that converts the undistorted frame to the HSV color space and applies empirically tuned hue-saturation-value thresholds to isolate the red droplet from the background. OpenCV contour detection extracts the droplet blob, and its centroid yields the pixel-space position. (ii) Four blue corner markers at known physical locations on the board are detected using the same masking procedure, and their pixel positions define a homography matrix that maps pixel coordinates to board-frame millimeter coordinates. (iii) Prior to the homography transformation, barrel and pincushion distortion are corrected using a fifth-order polynomial camera model obtained from the OCamCalib toolbox **[2]** calibrated on a checkerboard pattern, which fits the Scaramuzza omnidirectional polynomial model to a checkerboard calibration pattern and stores the resulting distortion coefficients in a calib_results.txt file. At runtime, each frame is undistorted using the stored polynomial before any position estimation is performed. (iv) The homography is computed once from the first valid corner detection and then locked for the remainder of the experiment, preventing false position estimates when a corner marker is temporarily occluded during board tilting. **Supplementary figure S2** shows processed frame. (v) Raw pixel detections are noisy and intermittently lost due to specular glare from the oil film. To provide the controller with a continuous position estimate, we apply a constant-velocity Kalman filter with four states $(x, y, v_x, v_y)$ and two measurements $(x, y)$. At each frame, the filter predicts the droplet position from its previous state and velocity, then blends this prediction with the camera measurement when available. When the droplet is not detected, the filter continues to predict using the last known velocity for up to 90 frames (approximately 3 seconds), bridging transient detection gaps. (vi) If the droplet remains undetected beyond 5 consecutive frames, motor commands are frozen at their last known values to prevent the policy from reacting to increasingly uncertain predictions. Remaining architecture includes data collection and training setup have been discussed in the main paper.

Lighting conditions must remain consistent across data collection sessions to ensure stable HSV thresholds; all experiments were conducted under fixed overhead illumination. The software stack runs on Robot Operating System 2 (ROS 2, Humble distribution). The camera driver, motor driver, position estimator, and policy runner each operate as independent ROS 2 nodes within a unified workspace (~/droplet_runner_ws/).

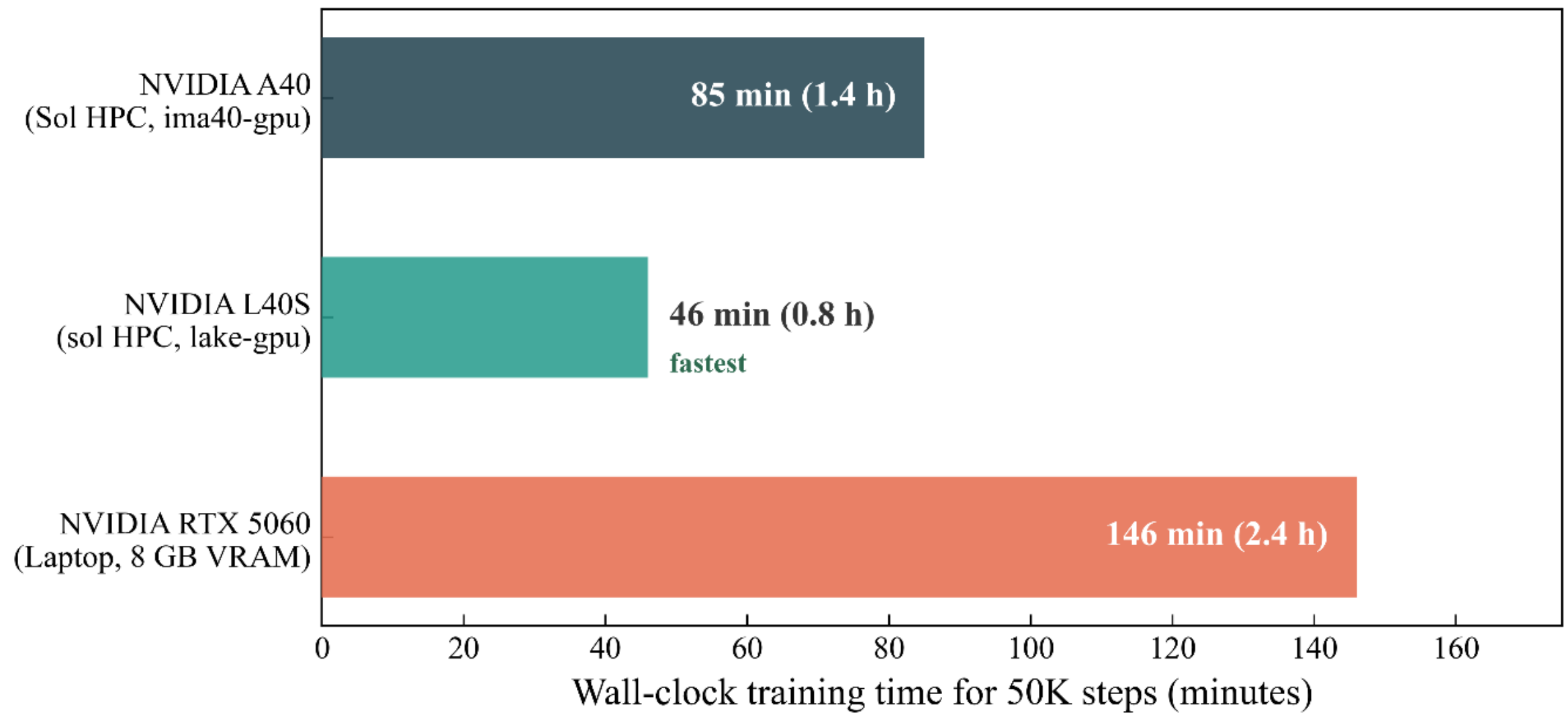

**Supplementary Figure S3 | GPU training benchmark for offline world model training.** Wall-clock time for 50,000 gradient update steps training the DreamerV3 world model from 100 offline episodes, measured on three GPU platforms: NVIDIA L40S (Lehigh University Sol HPC, lake-gpu partition; Ada Lovelace architecture, 48 GB GDDR6, 18,176 CUDA cores), NVIDIA A40 (Lehigh University Sol HPC, ima40-gpu partition; Ampere architecture, 48 GB GDDR6, 10,752 CUDA cores), and NVIDIA RTX 5060 (laptop; 8 GB GDDR7). The L40S completes training in 46 minutes, approximately two-fold faster than the A40 (85 minutes), consistent with the generational improvement from Ampere to Ada Lovelace in FP32 throughput and memory bandwidth (864 vs. 696 GB/s). RTX 5060 laptop GPU completes the same training in 146 minutes, approximately three-fold slower than the L40S but well within a single working session. Training was performed entirely offline, requiring no real-time GPU-hardware synchronization, which enables laboratories to collect data on low-cost CPU hardware and offload training to any available GPU resource, whether a shared institutional cluster or a personal laptop. The fact that a laptop GPU can train a policy achieving 95–100% navigation success across three distinct path geometries demonstrates that the approach imposes no prohibitive computational barrier, placing autonomous liquid droplet manipulation within reach of any laboratory equipped with standard computing hardware.

## 3. Control Unit

The control unit consists of two computers: a CPU-only laptop that runs the ROS2 control loop, data collection, and motor communication via the U2D2 adapter, and a GPU-equipped platform used exclusively for offline world-model training. The CPU laptop manages the entire experimental setup in real time, integrating data from the physical actuators and the imaging pipeline. Collected episodes are transferred to the GPU platform for offline training, and the resulting policy checkpoint is transferred back for evaluation. This decoupled architecture ensures that training latency does not interfere with real-time data collection and allows laboratories to use any available GPU resource for the training stage. We have also investigated the training time on different GPUs. **Supplementary figure S3** shows training time on three different GPU specifications.

## 4. Training the RL agent

Data collection costs remain a challenge when applying RL to physical micro-scale systems. During learning on the physical system, each episode requires manual placement of the droplet at the start position and periodic replenishment of the oil film, which degrades over several days of ambient exposure. Our system produces approximately 50 episodes per hour of operator time. We structure the training and deployment process into three stages (**Fig. 2d**).

### Stage 1: Random exploration

In the first stage, episodes are collected under a purely random policy with no learned checkpoint. At each time step, a random motor command is independently sampled for each motor from a zero-mean Gaussian distribution with standard deviation 80 mA. To prevent incoherent high-frequency vibration, the sampled command is blended with the previous command using a temporal smoothing filter (**Eq. 2**).

$$a_t = 0.7\, a_{t-1} + 0.3\, a_t^{\text{raw}} \quad (2)$$

where $a_t^{\text{raw}}$ is the freshly sampled random command and $a_t$ is the smoothed command sent to the motors, clipped to $\pm150$ mA. The 70% retention of the previous action ensures that the board tilts consistently in one direction for several time steps before gradually transitioning, producing actual droplet displacement that covers diverse regions of the board and path. This generates informative training data in which the world model can observe the causal relationship between sustained tilt commands and droplet motion. We collect 50 episodes under the random policy for each geometry. The choice of purely random exploration in Stage 1, rather than a simulation-based pretraining approach as employed in prior RL microrobotics work **[9]**, is deliberate. Computational modeling of the droplet-on-oil system would require resolving the Navier-Stokes equations coupled with dynamic contact-line models, oil film depletion, and meniscus deformation at geometric boundaries **[10-13],** which is highly sophisticated and computationally expensive. By collecting data directly on the physical system, the world model trains on the true dynamics from the outset, including all nonlinearities, surface heterogeneities, and unmodeled effects that a simulation would either omit or approximate poorly. Simulation-free approach makes the framework easily deployable in any laboratory setting without requiring physics modelling.

**Stage 2: World-model training**

The collected episodes are transferred from the control laptop (CPU) to a GPU platform for offline training. The training script converts the raw current values to the normalized action range $[-1, 1]$ by dividing by 150 (the maximum current), and performs 50,000 gradient updates on the world model, actor, and critic networks simultaneously. Each gradient step samples a batch of 16 sequences, each 64 time steps long, drawn randomly from the stored episodes.

**Stage 3: Evaluation and guided exploration**

After Stage 2 training, the resulting checkpoint (policy) is evaluated on the physical system. If the success rate is 100% or close, the training is considered complete. If the policy fails or achieves only partial success (less than 90%), which is typical after a single training cycle on complex geometries such as the L-shape and Arc, additional data is collected using the learned checkpoint in exploration mode. Instead of random commands, the DreamerV3 agent generates actions using its learned policy with internally added exploration noise:

$$a_t = \pi(o_t) + \epsilon_t, \qquad \epsilon_t \sim \mathcal{N}(0, \sigma^2) \quad (3)$$

where $\pi(o_t)$ is the deterministic policy output given the current observation and $\varepsilon_t$ is stochastic (exploration) noise. This produces episodes that are guided by the partially trained policy but include sufficient randomness to visit states the current policy has not yet mastered. The advantage over continued random exploration is that the learned policy reliably navigates segments it has already solved, reaching the challenging regions (the corner) more frequently and generating richer training data precisely where the world model needs improvement. A second batch is collected and transferred to the GPU platform (**Fig. 2d**). Training resumes from the existing weights rather than reinitializing. The final trained checkpoint is deployed on the control PC in evaluation mode. For each geometry, we evaluate the MBRL policy over 20 consecutive episodes. The droplet is manually placed at the start position before each episode, and the episode proceeds autonomously until termination. Success is defined as the droplet entering a 10 mm radius of the goal position. The number of training cycles required depends on the geometric complexity of the path. The I-shape converges in a single cycle of 50 random episodes, L-shape and Arc geometries require two cycles; the L-out and Arc-out require three cycles to reach a satisfactory (90%) success rate.

## 5. PID controller workflow

To quantify the advantage of the learned world model (MBRL) over classical reactive control, we implemented a discrete time proportional-integral-derivative (PID) waypoint-tracking controller operating on the same hardware, sensor pipeline, experimental setup, and actuation interface as the MBRL agent. The sign convention was determined empirically by applying step commands to each motor individually and observing the resulting droplet displacement direction. Motor 1, positive current displaces droplet in +x direction while Motor 2, positive current displaces droplet in -y direction. The sign parameters ($\sigma_1 = +1, \sigma_2 = -1$) were incorporated into the control law to ensure that positive positional error in each axis produces a motor command that drives the droplet toward the target. The gains were selected ($K_p = 4, K_i = 0.1, K_d = 2.0$) to balance three competing requirements: (i) sufficient control actions to overcome static friction and initiate droplet motion, (ii) derivative damping to prevent oscillatory overshoot on the low-friction oil film, and (iii) minimal integral contribution to avoid amplifying oscillations while still correcting persistent offsets. Multiple gain configurations were tested ($K_p \in \{3, 4, 6, 10\}, K_i \in \{0, 0.1, 0.3\}, K_d \in \{0.5, 2.0\}$); the reported values produced the best overall performance. PID control working algorithm has been shown in **Algorithm 2**. Identical criteria to the MBRL evaluation were applied for episode termination. An episode is classified as a success if the droplet reaches within 10 mm of the final waypoint, and as a failure if lateral deviation exceeds 30 mm or the step limit is reached.

The failure of PID controller on this system is not a subject of gain tuning but reflects a fundamental mismatch between reactive control and the physics of droplet transport on lubricated surfaces. The PID control law $u = K_p \cdot e + K_i \cdot \int e \, dt + K_d \cdot \dot{e}$ responds only to the current error and its history. On the oil-lubricated surface, the droplet exhibits low-friction sliding dynamics with significant momentum. Once set in motion, the droplet continues moving after the tilt is reversed, creating a phase lag between actuation and response. The derivative term provides some anticipatory braking, but only in response to observed error changes. It can't predict the droplet's future trajectory. The MBRL world model, by contrast, learns a latent dynamics model that captures this momentum and enables the actor to issue preemptive deceleration commands 5-10 steps before a turn. Moreover, when the droplet touches a wall, surface tension holds it against the wall. A constant tilt below a certain threshold isn't enough to move it, but (MBRL) rapid oscillation creates vibrational energy that overcomes the local hysteresis and droplet moves.

**Supplementary Note S2. Oscillation analysis methodology**

The raw motor-current signals contain multiple superimposed components (**Supplementary Figure S4a**), including high-frequency bang-bang switching from the reinforcement learning controller, slow navigational drift, sensor noise, and, critically, a rhythmic oscillation that the agent learned to employ for rocking the droplet free from corner obstacles. Isolating the frequency and time-varying amplitude of this specific depinning oscillation is essential for demonstrating that the agent discovered an emergent vibration-induced transport strategy paralleling the rectified-motion mechanism described by Daniel and Chaudhury [**4, 5**]. Standard spectral methods are inadequate for this task. A global fast Fourier transform (FFT) shows the existence of a spectral peak near 0.5–0.6 Hz but removes all temporal information. Short-time Fourier transform (STFT) recovers time resolution, but at approximately 0.6 Hz in a 28-second signal sampled at 20 Hz, the uncertainty principle demands analysis windows of at least 2–3 seconds to resolve the frequency. Continuous wavelets transform (CWT) redistributes this tradeoff but does not eliminate it: a Morlet wavelet at 0.6 Hz spreads over roughly 2 seconds, spreading the time axis comparably. We therefore adopted the Hilbert–Huang transform (HHT), which combines empirical mode decomposition (EMD) with the Hilbert transform to yield exact instantaneous frequency and amplitude at every timestep without any windowing tradeoff [**6**].

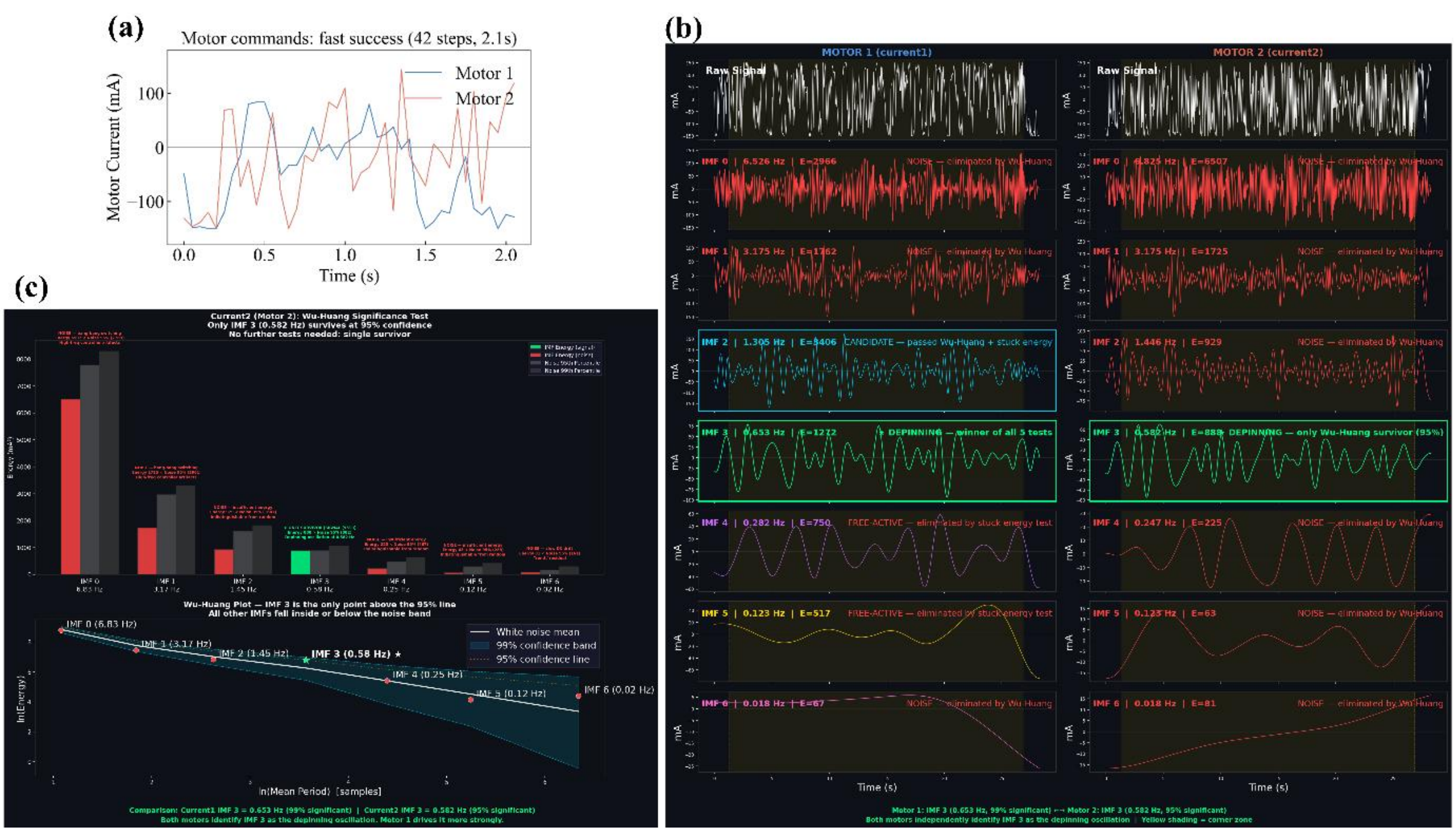

**Supplementary Figure S4 | Raw motor-current signals, full IMF decomposition, and Motor 2 Wu–Huang test. (a)** Representative raw Motor 1 (blue) and Motor 2 (red) current commands during the corner encounter, illustrating the superposition of bang-bang switching with a lower-frequency oscillatory component. **(b)** Complete EMD decomposition of Motor 1 (left) and Motor 2 (right) current signals into seven IMFs each (IMF 0–6), ordered from highest to lowest frequency. **(c)** Wu–Huang significance test for Motor 2. Top: bar chart comparing

each IMF's energy against the 95th and 99th noise percentiles. Only IMF 3 (0.582 Hz) exceeds the 95% threshold; all other IMFs are indistinguishable from noise. Bottom: the energy-period Wu-Huang plot confirming IMF 3 as the sole significant component (green star above the 95% confidence line).

EMD algorithm iteratively extracts oscillatory modes through an envelope-fitting sifting process: local maxima and minima of the signal are identified, cubic splines are fitted through each set of extrema to form upper and lower envelopes, the local mean (average of the two envelopes) is subtracted, and the procedure is repeated until the residual satisfies the intrinsic mode function (IMF) conditions, namely that the number of extrema and zero-crossings differ by at most one and that the local mean of the envelopes is everywhere approximately zero **[6]**. The identified IMF is then subtracted from the signal, and the entire sifting process is repeated on the residual to extract successively slower modes. For the Motor 1 current signal of a representative episode (567 samples, 28.35 s at 20 Hz), EMD produced seven IMFs spanning from approximately 6.5 Hz (fast controller switching, IMF 0) down to 0.02 Hz (slow DC drift, IMF 6), and a monotonic residual (**Supplementary Figure S4b**). Since EMD produces multiple IMFs, identifying which one corresponds to the physical depinning oscillation requires a principled selection procedure. We established this through a three-stage elimination protocol in which each stage employs independent data or logic (**Supplementary Figure S5**).

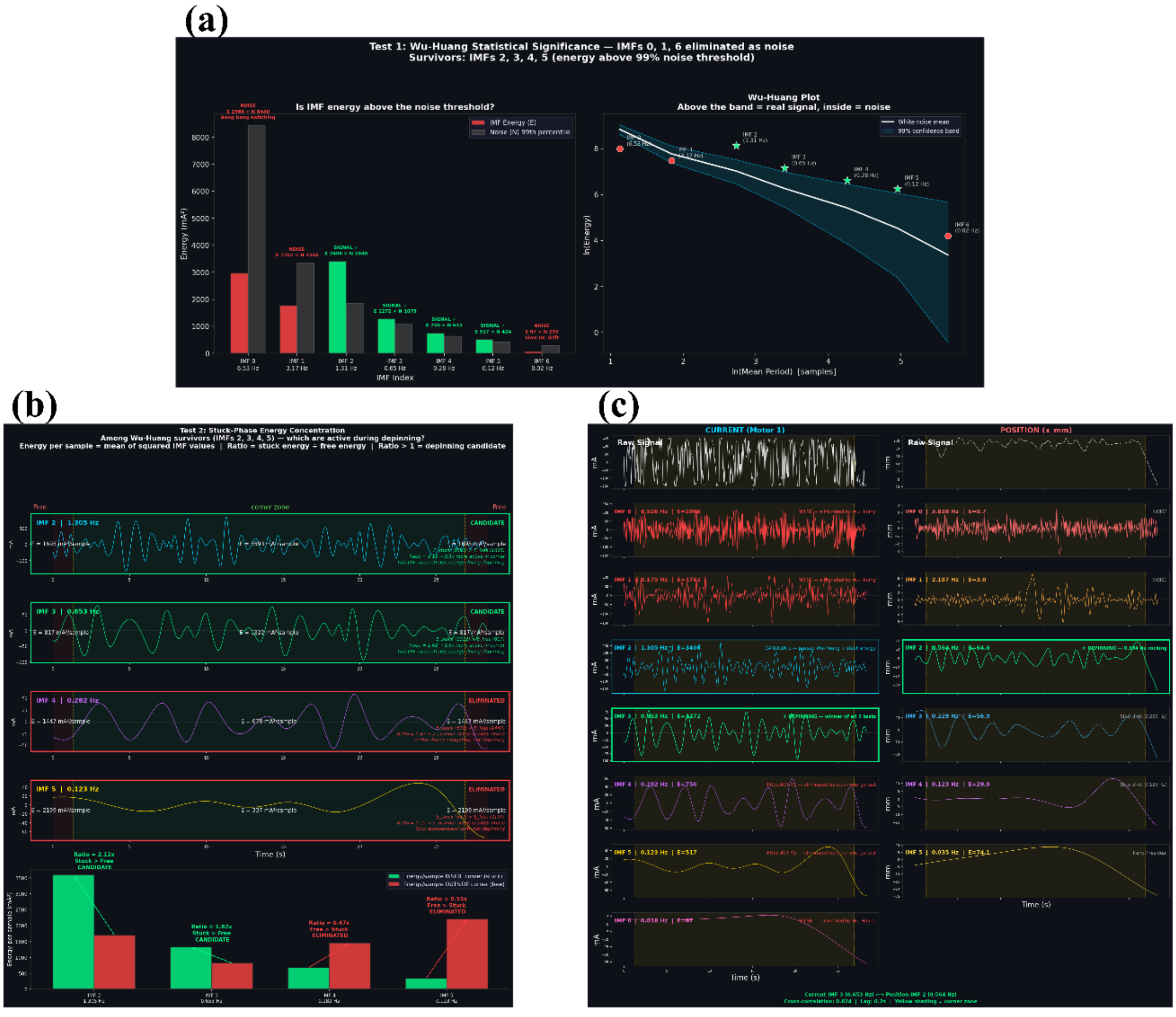

**Supplementary Figure S5 | EMD decomposition and three-stage IMF selection for Motor 1. (a)** Wu–Huang statistical significance test (Stage 1). Left: energy of each Motor 1 IMF (red) compared against the 99th percentile of 1,000 white-noise Monte Carlo realizations (grey). IMFs 0, 1 and 6 fall within the noise band and are eliminated; IMFs 2, 3, 4 and 5 contain statistically significant oscillatory content. Right: the corresponding Wu–Huang energy-period plot, where points above the 99% confidence band (cyan shading) denote significant IMFs (green stars). **(b)** Stuck-phase energy concentration test (Stage 2). For each surviving IMF, the time-domain waveform is shown with the corner zone shaded in green. Energy per sample inside versus outside the stuck phase is compared; IMFs 2 and 3 concentrate energy during the corner encounter and are retained as candidates, while IMFs 4 and 5 are eliminated. Bottom bar chart summarizes the stuck-versus-free energy ratios. **(c)** Cross-validation against droplet position (Stage 3). Left column shows Motor 1 current IMFs with elimination annotations. Right column shows independent EMD of the droplet x-position signal. The position signal's significant oscillatory component (IMF 2, 0.564 Hz) is closest in frequency to current IMF 3 (0.653 Hz, mismatch 0.088 Hz), confirming IMF 3 as the fundamental depinning mode.

***Stage 1: Wu–Huang significance testing***. Wu and Huang [**7**] showed that EMD applied to white noise acts as a dyadic filter bank, with the energy density of each IMF following a predictable chi-squared distribution. We exploited this property to construct a null-hypothesis test: 1,000 realizations of Gaussian white noise matching the length (567 samples) and variance of the motor-current signal were generated, EMD was applied to each realization, and the energy (variance) of each IMF was recorded to build a reference distribution at every timescale. Any real IMF whose energy fell within the 95th percentile of the corresponding noise distribution was deemed statistically indistinguishable from noise and eliminated. For Motor 1, this test removed IMFs 0, 1, and 6, leaving IMFs 2, 3, 4, and 5 as containing statistically significant oscillatory content (**Supplementary Figure S5a**). Importantly, this step involves no frequency-band preselection; significance is determined entirely by whether the observed energy exceeds what noise alone would produce at that timescale. For Motor 2, the test was even more decisive, only a single IMF (IMF 3, with a mean frequency of 0.582 Hz) exceeded the 95th percentile threshold, and all other IMFs were indistinguishable from noise (**Supplementary Figure S4c**). This result shows Motor 2's weaker contribution to the depinning oscillation and rendered subsequent stages unnecessary for that channel.

***Stage 2: Stuck-phase energy concentration***. The depinning oscillation must, by physical reasoning, be most active when the droplet is trapped at the corner obstacle. We defined the stuck phase as the time window during which the droplet remained within 30 mm of the corner waypoint. It is determined independently from the position tracking data (steps 26–537, approximately 1.3-26.9 s). For each surviving Motor 1 IMF, we computed the ratio of the mean-squared energy per sample inside the stuck phase to that outside it. **Supplementary Figure S5b** shows that IMFs 2 and 3 have ratios of 2.12 and 1.62, respectively, indicating that they concentrate energy during the corner encounter. IMFs 4 and 5, with ratios of 0.47 and 0.15, are more energetic during free navigation and therefore cannot represent the depinning oscillation. This stage reduced the Motor 1 candidate set to two IMFs.

***Stage 3: Cross-validation against droplet position***. To disambiguate the two remaining candidates, we performed a fully independent EMD on the droplet's x-position time series (obtained from the vision-based tracking system, with no dependence on the motor-current data). Applying the same Wu-Huang significance test to the position IMFs identified a significant oscillatory component at 0.564 Hz, representing the physical rocking frequency. Among the two surviving current IMFs, IMF 3 (a mean frequency 0.653 Hz) was closest to this position-derived frequency, with a mismatch of only 0.088 Hz, as shown in **Supplementary Figure S5c**. IMF 2 (1.305 Hz), at approximately double the physical frequency.

With the depinning IMF identified (IMF 3 for both motors), the Hilbert transform was applied to obtain the analytic signal $z(t) = m(t) + i\,\mathcal{H}[m(t)] = A(t)\,e^{i\phi(t)}$, where $\mathcal{H}$ denotes the Hilbert transform [**8**]. The instantaneous amplitude $A(t) = |\,z(t)\,|$ quantifies the vibration strength (in mA) at each timestep, and the instantaneous frequency $f(t) = \frac{1}{2\pi}\,\mathrm{d}\phi/\mathrm{d}t$ is computed from the unwrapped phase via central finite differences.

**Supplementary Note S3. Fabrication of the 3D-printed PLA board**

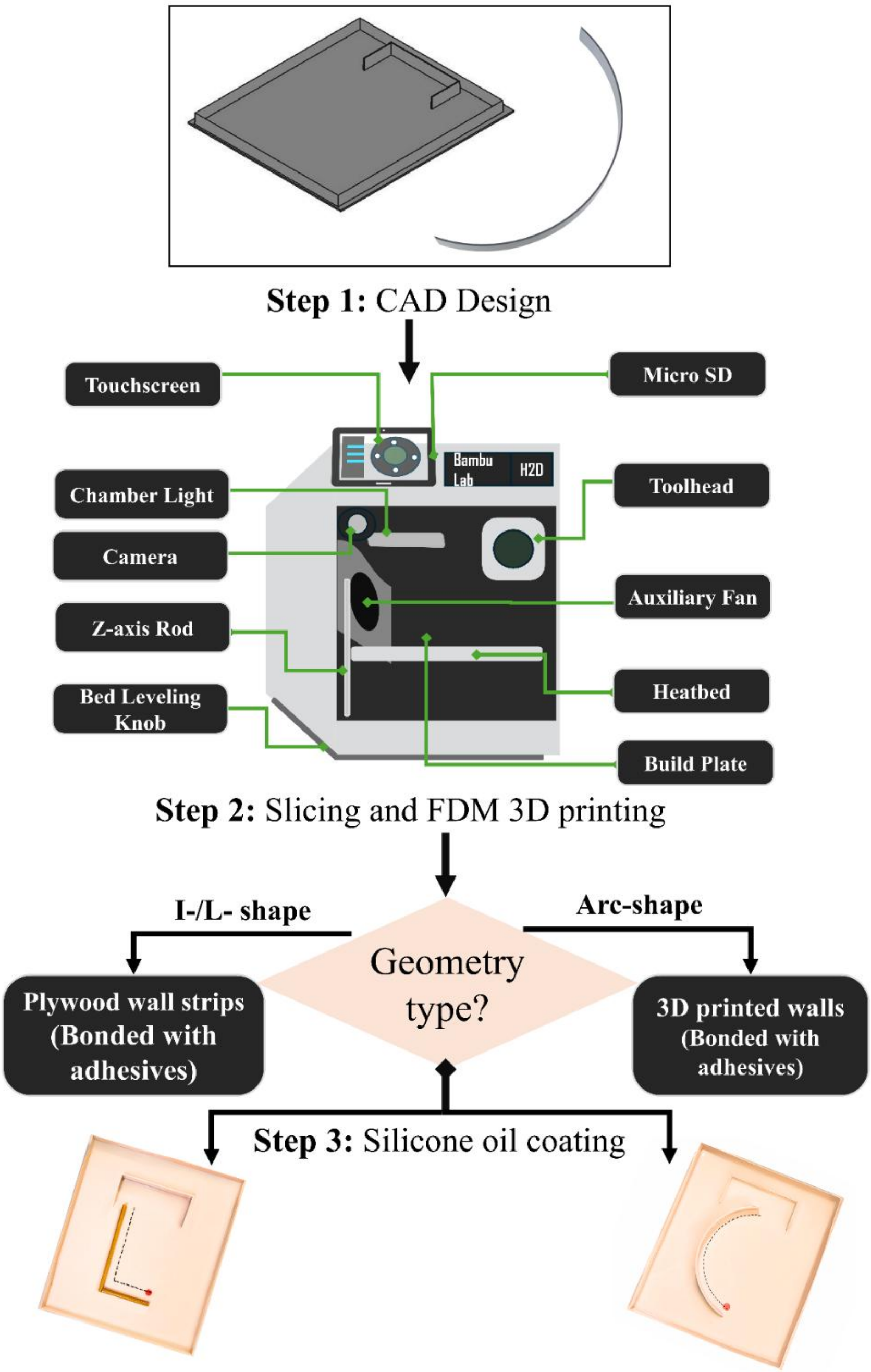

**Supplementary Figure S6 | Fabrication workflow for navigation surface boards.** Step 1: the flat base plate and arc are designed as parametric CAD models in SolidWorks and exported as STL files. Step 2: G-code toolpaths are generated in Bambu Studio and printed. A geometry-dependent branching determines wall construction: plywood strips bonded with adhesive for the I-shape and L-shape channels, and 3D-printed PLA curved segments for the Arc geometries. Step 3: the assembled board is coated with AR 20 silicone oil. Bottom panels show representative finished boards for the L-shape (left) and Arc-shape (right) geometries with red dyed water droplet at the start point.

The flat navigation boards used in this study were fabricated by fused deposition modeling (FDM) on a Bambu Lab H2D 3D printer (Bambu Lab) using Polymaker PLA filament (Polymaker) with a 0.4 mm nozzle. Board geometries were designed in SolidWorks as parametric CAD models with outer dimensions matched to the BRIO Labyrinth frame (approximately 265 mm × 225 mm). The CAD models were exported as STL files and sliced using Bambu Studio slicer software to generate G-code toolpaths, which were uploaded to the printer via Wi-Fi. For the I-shape and L-shape geometries, the base board was printed as a flat PLA plate; channel walls were formed by bonding thin plywood strips (approximately 8 mm height) onto the base with adhesive, and the channel centerline was marked with a fine-tip permanent marker to provide a visual reference during waypoint calibration. For the Arc and Arc-out geometries, the curved wall segments (radius of curvature = 80 mm, sweep angle 150°) were designed as separate parametric parts in SolidWorks, 3D-printed from PLA, and bonded to the flat base board with adhesive. After assembly, the navigation surface was coated with a thin film of AR 20 silicone oil (viscosity 20 mPa-s at 25 °C; Sigma-Aldrich) dispensed from a pipette and spread uniformly across the board surface. The complete fabrication workflow is shown in **Supplementary Figure S6**.

**Hardware components**

| **Component** | **Specification** | **Role** |
|---|---|---|
| Camera | See3CAM 24CUG, 1280x720 | Overhead droplet tracking via HSV color detection |
| Motors | Dynamixel XL330-M077 x2, U2D2 controller | Two-axis board tilt actuation (current control mode) |
| Board | Beige PLA 3D-printed surface, 266mm x 241mm | Oil-coated labyrinth with painted black path for droplet navigation |
| Droplet | Red-dyed water on silicone oil film | The object being navigated |
| Corner Markers | 4 blue dots on board corners | Homography reference for pixel-to-mm conversion |
| Laptop 1 | HP laptop, Ubuntu 22.04, no GPU | ROS2 control, data collection, policy inference |
| Laptop 2 / Sol HPC | RTX 5060 8GB / A40 48GB | Offline DreamerV3 training |

**Supplementary Table S1 | Hardware components and their use**

**Algorithms and hyperparameters used by the RL agent.**

| **Parameter** | **Details** |
|---|---|
| Algorithm | DreamerV3 (Model-Based Reinforcement Learning) |
| World model (RSSM) | GRU deterministic state: 512 units;<br>Stochastic state: 32 × 32 categorical; |
| | Learning rate: $1 \times 10^{-4}$ (Adam, grad clip = 1000) |
| Actor (policy) | 2-layer MLP, 512 units, SiLU, LayerNorm;<br>Distribution: Normal ($\sigma \in [0.1,1.0]$);<br>Gradient: backpropagation;<br>Learning rate: $3 \times 10^{-5}$ (Adam, grad clip = 100) |
| Critic (value) | 2-layer MLP, 512 units, SiLU, LayerNorm;<br>Distribution: symlog discrete (255 bins);<br>Slow EMA fraction: 0.02;<br>Learning rate: $3 \times 10^{-5}$ (Adam, grad clip = 100) |
| Parameter count | World model: 17.3M;<br>Actor: 1.05M;<br>Critic: 1.18M;<br>Total: 19.5M |
| Input | $64 \times 64 \times 3$ RGB image +<br>14D vector $[x, y, \alpha, \beta, 10$ lookahead offsets$]$ |
| Output | 2D continuous action $\in [-1,1]$, scaled by 150 mA at deployment |
| Batch size | 16 |
| Sequence length | 64 |
| Gradient updates | 50,000 |
| Discount ($\gamma$) | 0.997 (horizon = 333) |
| Lambda (GAE) | 0.95 |
| Training ratio | Training ratio |
| KL loss scales | Dynamics: 0.5, Representation: 0.1 |
| Action normalization | Raw motor current / 150 mA |
| Actor entropy coefficient | $3 \times 10^{-4}$ |
| Replay buffer | Uniform, capacity $1 \times 10^{6}$ |

**Supplementary Table S2 |** Architecture, hyperparameters, inputs, and outputs of the DreamerV3 MBRL agent. All other values follow the defaults of Hafner et al **[3]**.

## Other Supplementary Information: Extended Data Figures 1 and 2

### Extended Data Figures

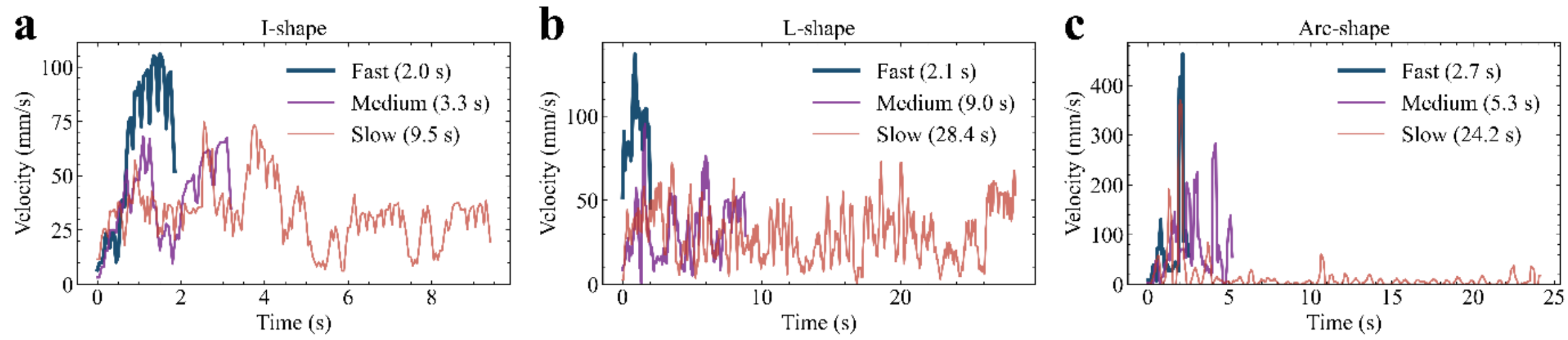

**Extended Data Figure 1 | Velocity profile across three path geometries.** Instantaneous droplet velocity as a function of time for three representative episodes (fast, medium, slow by completion time) on the **a,** I-shape, **b,** L-shape, and **c,** Arc geometries. Fast episodes (dark blue) sustain higher mean velocities throughout their trajectories, indicating that completion speed is governed by the amplitude of effective momentum transfer per oscillation cycle.

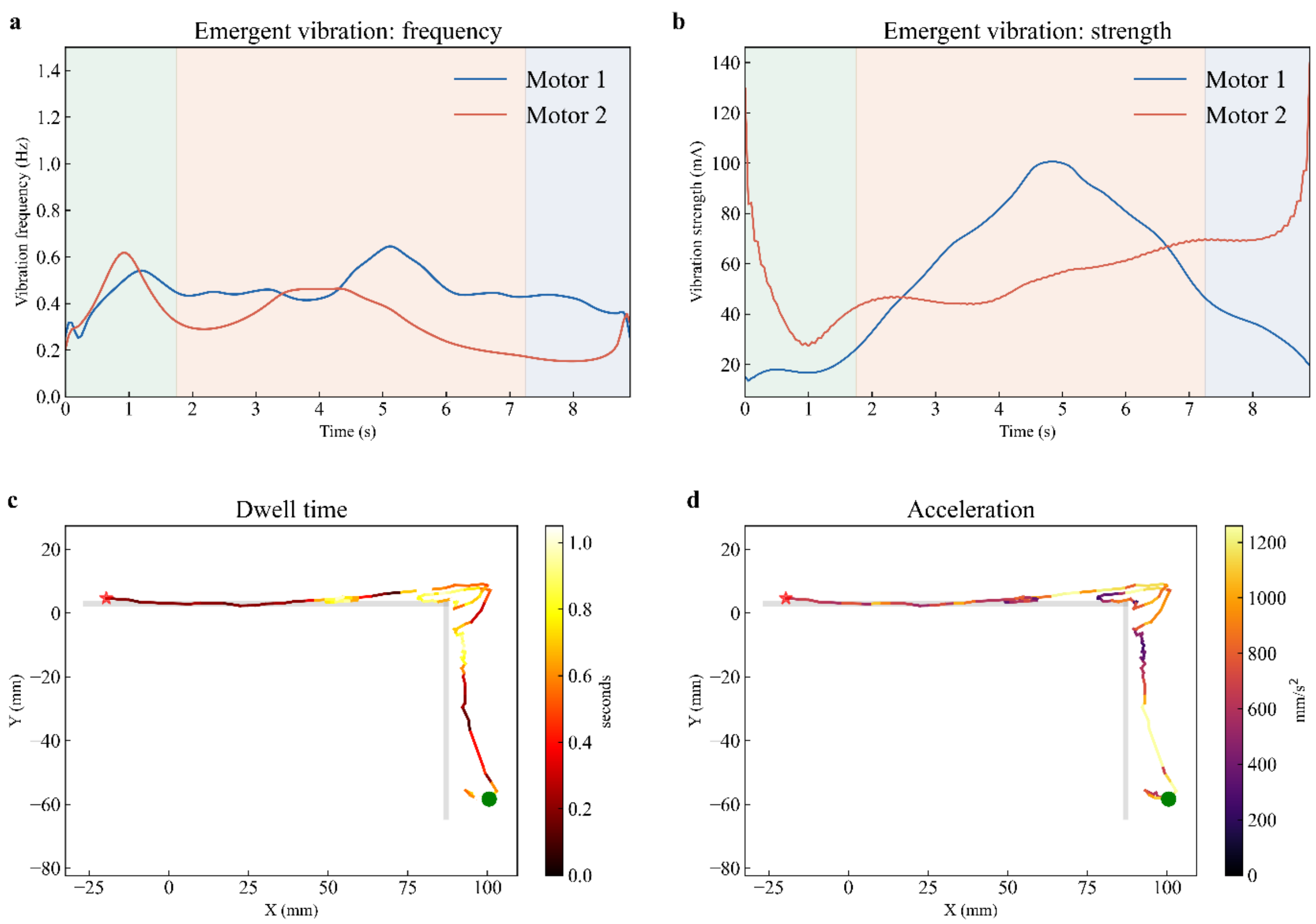

**Extended Data Figure 2 | Phase-resolved vibration analysis of a 179-step (8.95 s, L-shape) medium-duration episode. a,** Instantaneous vibration frequency and **b,** vibration strength of the motor current commands, computed via Hilbert transform with the same methodology as **Fig. 4e, f**. Three navigation phases are identified: approach (green shading), corner (pink shading), and escape (blue shading). In contrast to the 567-step episode (**Fig. 4e, f**), where the corner phase spans approximately 23 seconds and contains multiple depinning attempts, the 179-step episode resolves the corner in a single escalation cycle. Motor 1 amplitude rises from

approximately 10 mA during the approach phase to a peak of 100 mA at the corner, a tenfold increase, before falling in the escape phase. The clean ramp-up and recovery profile shows that the repeated oscillatory peaks observed in the longer 567-step episode (**Fig. 4f**) are not stochastic noise but discrete depinning events, each individually resembling the single-cycle pattern seen here. The approach and escape phases are of comparable duration (each approximately 2 seconds) and exhibit similar low-amplitude baselines. **c,** Trajectory of the 179-step episode colored by local dwell time. The bright cluster at the L-corner (dwell time approaching 1 s) confirms prolonged stagnation at the corner. **d,** The same trajectory colored by instantaneous acceleration magnitude. Acceleration exceeding 1200 mm/s² concentrates at the corner region, coinciding spatially with the dwell time maximum, indicating that the agent applies its most aggressive oscillatory forcing precisely where the droplet encounters the greatest resistance.

**Table 1 |** Sign change (SC) rate comparison between L- and L-out- shape geometries

| **Metric** | **L- shape (N = 19/20 ep.)** | **L-out- shape (N= 18/20 ep.)** | **Change** |
|---|---|---|---|
| Motor#1 SC rate | $27 \pm 6\%$ | $21 \pm 4\%$ | −6% |
| Motor#2 SC rate | $42\pm10\%$ | $19 \pm 4\%$ | −23% |

**Supplementary movies S1 to S15**

An archived version of all supplementary movies has also been uploaded to Google Drive:

https://drive.google.com/drive/folders/1ewK2dxWxfzjd4kk6Df4cqKE3-3xCbOgf?usp=sharing